\pdfoutput=1
\documentclass[11pt]{article}
\usepackage[preprint]{acl}
\usepackage{times}
\usepackage{latexsym}
\usepackage{hyperref}
\usepackage[T1]{fontenc}
\usepackage[utf8]{inputenc}
\usepackage{microtype}
\usepackage{inconsolata}
\usepackage{graphicx}
\usepackage{booktabs}
\usepackage{multirow}
\usepackage[most]{tcolorbox}
\usepackage{multicol}

\title{Smaller Language Models Are Better Instruction Evolvers}
\author{
 \textbf{Tingfeng Hui\textsuperscript{1}\thanks{denotes equal contribution. Work done during Hui's internship at BAAI.}},
 \textbf{Lulu Zhao\textsuperscript{2}\footnotemark[1]},
  \textbf{Guanting Dong\textsuperscript{3}},
 \textbf{Yaqi Zhang\textsuperscript{1}},
 \textbf{Hua Zhou\textsuperscript{2}},
 \textbf{Sen Su\textsuperscript{1}\thanks{The corresponding author.}}
\\
\\
 \textsuperscript{1}Beijing University of Posts and Telecommunications, Beijing, China \\
 \textsuperscript{2}Beijing Academy of Artificial Intelligence, BAAI, Beijing, China \\
 \textsuperscript{3}Renmin University of China, Beijing, China \\
 \textsuperscript{1}\texttt{(huitingfeng,zhangyaqi2021)@bupt.edu.cn} \\
 \textsuperscript{2}\texttt{llzhao@baai.ac.cn}
}

\begin{document}
\maketitle
\begin{abstract}
Instruction tuning has been widely used to unleash the complete potential of large language models.  Notably, complex and diverse instructions are of significant importance as they can effectively align models with various downstream tasks. However, current approaches to constructing large-scale instructions predominantly favour powerful models such as GPT-4 or those with over 70 billion parameters, under the empirical presumption that such larger language models (LLMs) inherently possess enhanced capabilities. In this study, we question this prevalent assumption and conduct an in-depth exploration into the potential of smaller language models (SLMs) in the context of instruction evolution. Extensive experiments across three scenarios of instruction evolution reveal that smaller language models (SLMs) can synthesize more effective instructions than LLMs. Further analysis demonstrates that SLMs possess a broader output space during instruction evolution, resulting in more complex and diverse variants. We also observe that the existing metrics fail to focus on the impact of the instructions. Thus, we propose Instruction Complex-Aware IFD (IC-IFD), which introduces instruction complexity in the original IFD score to evaluate the effectiveness of instruction data more accurately. 
Our source code is available at: 
\href{https://github.com/HypherX/Evolution-Analysis}{https://github.com/HypherX/Evolution-Analysis}
\end{abstract}

\section{Introduction}
Large Language Models (LLMs) have demonstrated exceptional performance in various NLP tasks and are widely integrated into a variety of applications, represented by ChatGPT and Copilot~\citep{DBLP:conf/nips/Ouyang0JAWMZASR22, DBLP:journals/corr/abs-2303-08774, DBLP:journals/corr/abs-2407-21783}. A key factor in unleashing the full potential of these models is high-quality instruction tuning data, which plays a crucial role in post-training and enhances their effectiveness as AI assistants. In particular, incorporating more complex and diverse instructions allows models to better align with different domains and tasks, boosting their performance in a variety of downstream applications~\citep{DBLP:journals/corr/abs-2308-10792}. However, generating such diverse instructions remains time-consuming and labor intensive~\citep{DBLP:conf/iclr/ZhengC0LZW00LXG24, DBLP:conf/iclr/Zhao0HC0D24, DBLP:conf/iclr/0131Z00H24}, which undoubtedly presents a significant challenge for the automated and scalable alignment of LLMs. Recently, a series of efforts utilizing LLMs for automatic instruction evolution have garnered sustained attention from the community. Specifically, foundational work like Self-Instruct~\citep{DBLP:conf/acl/WangKMLSKH23} begins with a small set of seed instructions and uses a powerful supervision model to obtain a large number of synthetic instructions. Furthermore, Evol-Instruct~\citep{DBLP:conf/iclr/XuSZG0FTLJ24} refines and evolves existing instructions to produce more complex variants. 

However, previous studies mainly favour strong LLMs like GPT-4 or those with more than 70 billion parameters to synthesize instructions, empirically assuming that larger language models inherently have superior instruction evolution capabilities. But is this really the case? Recently, \citet{xu2024strongermodelsstrongerteachers} propose the Larger Models' Paradox, which points out that larger models do not necessarily lead to better performance when generating responses, but it overlooks the analysis of instructions. We propose that smaller language models, which require less computational demand and have lower instruction following capabilities, may provide a more efficient and effective alternative for evolving more complex and diverse instructions. To gain insight into this, we investigate the differences between smaller language models (SLMs) and larger language models (LLMs) in generating high-quality instructions. Specifically, given a set of base models and seed instructions, we are particularly interested in the following research question:

\textit{RQ1: Do SLMs Perform Better than LLMs in Evolving Instructions?}

In response to this, we conduct comprehensive experiments across three distinct instruction evolution scenarios: Evol-Instruct, AutoIF~\citep{DBLP:journals/corr/abs-2406-13542}, and Auto Evol-Instruct~\citep{DBLP:conf/emnlp/ZengXZLC24}. In these experiments, we use small ($\sim$8B) and large ($\sim$70B) models from the Llama-3.1 and Qwen-2 families to evolve and synthesize new instructions, while also fine-tuning various backbone models. The experimental results across all three scenarios consistently indicate that larger, more powerful LLMs do not outperform SLMs in evolving effective instructions. More interestingly, SLMs even demonstrate the capability to evolve more complex and diverse instructions. To further investigate why more powerful LLMs perform worse than SLMs in generating new instructions, we subsequently pose the second research question.

\textit{RQ2: Why do SLMs Outerperform LLMs in Evolving Instructions?}

To better understand why more powerful LLMs underperform compared to SLMs in evolving instructions, we compare the top-1 token probabilities of both models during the synthetic of instructions. Our findings demonstrate that LLMs, due to their superior instruction following capabilities, tend to generate a higher proportion of high-probability top-1 tokens when evolving new instructions. This overconfidence in token generation results in a narrower output space. In contrast, SLMs can generate a wider variety of tokens, leading to more complex and diverse instructions. To further investigate what kind of instruction data is effective, we propose the third research question.

\textit{RQ3: How Do We Determine Whether An Instruction is Effective without Instruction Tuning?}

Evaluations that do not require instruction tuning can more efficiently assess instruction data. Recent such evaluations often fail to account for the impact of the instructions themselves. For instance, reward models~\citep{DBLP:journals/corr/abs-2403-17297} are commonly used to assess the quality of responses generated based on a given instruction, yet they tend to overlook the quality of the instruction itself. Similarly, while the IFD score~\citep{DBLP:conf/naacl/LiZLCC0W0024} measures the influence of instructions on response generation, it neglects the effect of the instruction's inherent complexity. We introduce the Instruction Complex-Aware IFD (IC-IFD) score, which incorporates the difficulty of the instruction as a penalty term in the original IFD. We conduct extensive filtering instruction data experiments, and the results demonstrate that the IC-IFD score provides a more accurate assessment of instruction data, particularly in scenarios where the instructions exhibit higher complexity levels. In summary, our key contributions are as follows:

(1) To the best of our knowledge, we are the first to comprehensively explore the performance discrepancies between SLMs and LLMs in synthesizing instructions.

(2) Extensive experimental results demonstrate that SLMs have a broader output space, leading to evolving more complex and diverse instructions.

(3) We propose the IC-IFD score, which introduces the difficulty of the instruction as a penalty term. Comprehensive experiments show that IC-IFD can more accurately assess the effectiveness of instruction data without instruction tuning.

\begin{table*}[h]
    \centering
    \small
        \begin{tabular}{lcccccccccc}
            \toprule
            \multirow{2}{*}{\textbf{Model}} & \multicolumn{4}{c}{\textbf{Instruction Following (IFEval)}} & & \multicolumn{2}{c}{\textbf{Math Reasoning}} & & \multicolumn{2}{c}{\textbf{Code Generation}} \\
            \cmidrule{2-11}
            & \textbf{Pr.(S)} & \textbf{In.(S)} & \textbf{Pr.(L)} & \textbf{In.(L)} & & \textbf{GSM8K} & \textbf{MATH} & & \textbf{HumanEval} & \textbf{MBPP} \\
            \midrule
            \multicolumn{11}{c}{\textit{Supervised Model: Llama-3.1-70B-Instruct}}  \\
            Mistral-7B-v0.3 & 19.59 & 31.77 & 22.74 & 34.65 & \vrule & 33.89 & \textbf{3.16} & \vrule & 24.39 & 6.00 \\
            DeepSeek-7B & 36.23 & \textbf{48.20} & 41.04 & 52.52 & \vrule & \textbf{48.07} & 2.96 & \vrule & 28.66 & 33.00 \\
            Llama-3.2-3B & 40.11 & 50.84 & 43.81 & 54.43 & \vrule & 53.75 & 6.60 & \vrule & 35.98 & \textbf{36.00} \\
            Llama-3-8B & 33.83 & 46.28 & 36.41 & 49.28 & \vrule & 63.00 & 7.62 & \vrule & 43.90 & 36.20 \\
            Llama-3.1-8B & 34.57 & 46.04 & 38.81 & 50.48 & \vrule & 64.22 & 11.32 & \vrule & \textbf{51.22} & 40.60 \\
            InternLM-2-7B & 40.85 & 53.48 & 44.54 & 56.95 & \vrule & \textbf{68.31} & 19.50 & \vrule & 56.10 & 40.40 \\
            \midrule
            \multicolumn{11}{c}{\textit{Supervised Model: Llama-3.1-8B-Instruct}} \\
            Mistral-7B-v0.3 & \textbf{24.40} & \textbf{35.01} & \textbf{26.25} & \textbf{37.53} & \vrule & \textbf{40.18} & 2.84 & \vrule & \textbf{29.27} & \textbf{19.60} \\
            DeepSeek-7B & \textbf{36.60} & 48.08 & \textbf{41.77} & \textbf{53.12} & \vrule & 47.92 & \textbf{3.56} & \vrule & \textbf{34.76} & \textbf{33.80} \\
            Llama-3.2-3B & \textbf{41.59} & \textbf{53.48} & \textbf{45.66} & \textbf{57.07} & \vrule & \textbf{55.12} & \textbf{7.32} & \vrule & \textbf{39.02} & 32.80 \\
            Llama-3-8B & \textbf{35.49} & \textbf{47.00} & \textbf{39.56} & \textbf{50.72} & \vrule & \textbf{63.38} & \textbf{11.44} & \vrule & \textbf{48.17} & \textbf{37.60} \\
            Llama-3.1-8B & \textbf{38.45} & \textbf{50.96} & \textbf{43.81} & \textbf{55.28} & \vrule & \textbf{67.10} & \textbf{13.12} & \vrule & 48.78 & \textbf{41.60} \\
            InternLM-2-7B & \textbf{43.07} & \textbf{54.80} & \textbf{47.32} & \textbf{58.39} & \vrule & 68.08 & \textbf{20.32} & \vrule & \textbf{57.93} & \textbf{40.80} \\
            \bottomrule
        \end{tabular}
    \caption{Comparison of performance with Llama-3.1-8B-Instruct and Llama-3.1-70B-Instruct as supervised models under Evol-Instruct scenario.}
    \label{tab:llama_result}
\end{table*}

\section{Preliminaries}
\paragraph{(Auto) Evol-Instruct.} The goal of (Auto) Evol Instruct is to refine original instructions by using artificially designed or LLM-generated evolutionary trajectories, thereby increasing their complexity and fostering the development of a more capable model. Formally, given an instruction evolution model $\Theta_{e}$, a response generation model $\Theta_{r}$, and an original instruction dataset $\mathcal{D}=\{(\mathcal{I}_i,\mathcal{R}_i)\}_{i=1}^{n}$, where $\mathcal{I}$ and $\mathcal{R}$ are instructions and responses and $n$ represents the data size, we employ either artificially designed methods or the $\Theta_{e}$-generated evolutionary trajectory $\mathcal{T}$ to obtain more complex and diverse evolutionary dataset $D_{evol}=\{(\mathcal{I}_{ei}=\Theta_{e}(\mathcal{I}_{i}|\mathcal{T}),\mathcal{R}_{ei}=\Theta_{r}(\mathcal{R}|\mathcal{I}_{ei}))\}_{i=1}^{n}$.

\begin{table*}[ht]
    \centering
    \small
        \begin{tabular}{lcccccccccc}
            \toprule
            \multirow{2}{*}{\textbf{Model}} & \multicolumn{4}{c}{\textbf{Instruction Following (IFEval)}} & & \multicolumn{2}{c}{\textbf{Math Reasoning}} & & \multicolumn{2}{c}{\textbf{Code Generation}} \\
            \cmidrule{2-11}
            & \textbf{Pr.(S)} & \textbf{In.(S)} & \textbf{Pr.(L)} & \textbf{In.(L)} & & \textbf{GSM8K} & \textbf{MATH} & & \textbf{HumanEval} & \textbf{MBPP} \\
            \midrule
            \multicolumn{11}{c}{\textit{Supervised Model: Qwen-2-72B-Instruct}} \\
            Mistral-7B-v0.3 & 20.15 & 30.94 & 23.84 & 34.41 & \vrule & 46.93 & \textbf{3.26} & \vrule & 32.32 & 1.80 \\
            DeepSeek-7B & 35.67 & 47.12 & \textbf{39.56} & 50.84 & \vrule & 44.81 & 2.76 & \vrule & \textbf{36.59} & \textbf{34.00} \\
            Llama-3.2-3B & 39.74 & 51.44 & 43.99 & 55.40 & \vrule & 53.83 & \textbf{7.40} & \vrule & 38.41 & 31.00 \\
            Llama-3-8B & 34.75 & 45.80 & 37.71 & 48.92 & \vrule & 63.76 & \textbf{10.06} & \vrule & 43.90 & 35.40 \\
            Llama-3.1-8B & \textbf{36.41} & \textbf{47.60} & 39.00 & 50.60 & \vrule & 65.43 & 10.84 & \vrule & \textbf{48.17} & 38.40 \\
            InternLM-2-7B & 41.96 & 53.60 & 43.99 & 55.64 & \vrule & 65.28 & 17.96 & \vrule & 56.71 & 40.60 \\
            \midrule
            \multicolumn{11}{c}{\textit{Supervised Model: Qwen-2-7B-Instruct}} \\
            Mistral-7B-v0.3 & \textbf{25.32} & \textbf{37.17} & \textbf{29.76} & \textbf{41.01} & \vrule & \textbf{47.31} & 2.20 & \vrule & \textbf{32.93} & \textbf{12.00} \\
            DeepSeek-7B & \textbf{36.41} & \textbf{48.56} & 39.37 & \textbf{51.32} & \vrule & \textbf{48.07} & \textbf{3.82} & \vrule & 35.37 & 33.20 \\
            Llama-3.2-3B & \textbf{43.81} & \textbf{55.16} & \textbf{47.87} & \textbf{58.27} & \vrule & \textbf{56.56} & 7.18 & \vrule & \textbf{39.63} & \textbf{31.40} \\
            Llama-3-8B & \textbf{38.92} & \textbf{48.33} & \textbf{43.81} & \textbf{52.19} & \vrule & \textbf{63.91} & 8.66 & \vrule & \textbf{45.73} & \textbf{38.40} \\
            Llama-3.1-8B & 34.75 & 45.80 & \textbf{39.93} & \textbf{51.08} & \vrule & \textbf{68.76} & \textbf{14.02} & \vrule & 46.34 & \textbf{38.60} \\
            InternLM-2-7B & \textbf{44.12} & \textbf{55.16} & \textbf{48.62} & \textbf{58.73} & \vrule & \textbf{66.87} & \textbf{19.60} & \vrule & \textbf{58.54} & \textbf{41.40} \\
            \bottomrule
        \end{tabular}
    \caption{Comparison of performance with Qwen-2-7B-Instruct and Qwen-2-72B-Instruct as supervised models under Evol-Instruct scenario.}
    \label{tab:qwen_result}
\end{table*}

\paragraph{AutoIF.} The goal of AutoIF is to automatically construct large-scale and reliable instructions from a small set of seed instructions (which can also be seen as constraints) to improve instruction following ability. In this paper, we only utilize the first several steps of AutoIF. Specifically, given a small set of seed instructions $\mathcal{I}_{s}$, we first prompt the supervised model $\Theta$ to construct a large number of verifiable instructions $\mathcal{I}_{new}$ based on $\mathcal{I}_{s}$. Subsequently, we prompt $\Theta$ to generate the corresponding verification functions $f$ and test cases $c$ for $\mathcal{I} = \{\mathcal{I}_{s}, \mathcal{I}_{new}\}$. Finally, cross-validation is performed to obtain the final scalable and reliable instructions $\mathcal{I}_{final} = \{\mathcal{I} | f(\mathcal{I}, c) = True\}$.

\section{RQ1: Do SLMs Perform Better than LLMs in Evolving Instructions?}
In this section, we investigate the potential of SLMs in evolving complex and diverse instructions across three distinct scenarios: Evol-Instruct, AutoIF, and Auto Evol-Instruct. Through a series of comprehensive experiments and analyses, we attempt to answer the questions raised in RQ1. For clarity, we will refer to the instruction data evolved by SLMs and LLMs as \textsc{SLM-Inst} and \textsc{LLM-Inst}. The implementation details for the three scenarios, as well as our experimental hyperparameters, can be found in Appendix~\ref{sec:more_details}.

\subsection{Evol-Instruct Scenario}
\label{sec:evol_instruct}
In this section, we primarily focus on \textit{whether SLMs can evolve more complex and challenging instruction data compared to LLMs.}

\paragraph{Seed Datasets.} Following~\citep{DBLP:conf/iclr/XuSZG0FTLJ24, DBLP:conf/emnlp/ZengXZLC24}, we utilize the following seed datasets for instruction following, mathematical reasoning and code generation: (1) Alpaca~\citep{alpaca}, (2) GSM8K Train~\citep{DBLP:journals/corr/abs-2110-14168}, and (3) Code Alpaca~\citep{codealpaca}. More detailed information can be found in Appendix~\ref{sec:seed_data}.

\paragraph{Evaluation Benchmarks and Metrics.} We use IFEval~\citep{DBLP:journals/corr/abs-2311-07911} to assess instruction following capability, GSM8K and MATH~\citep{DBLP:conf/nips/HendrycksBKABTS21} to evaluate mathematical reasoning ability, and HumanEval~\citep{DBLP:journals/corr/abs-2107-03374} and MBPP~\citep{DBLP:journals/corr/abs-2108-07732} to assess code generation performance. For detailed information, please refer to Appendix~\ref{sec:evaluations}.

\begin{figure*}[ht]  
    \centering
    \includegraphics[width=\textwidth]{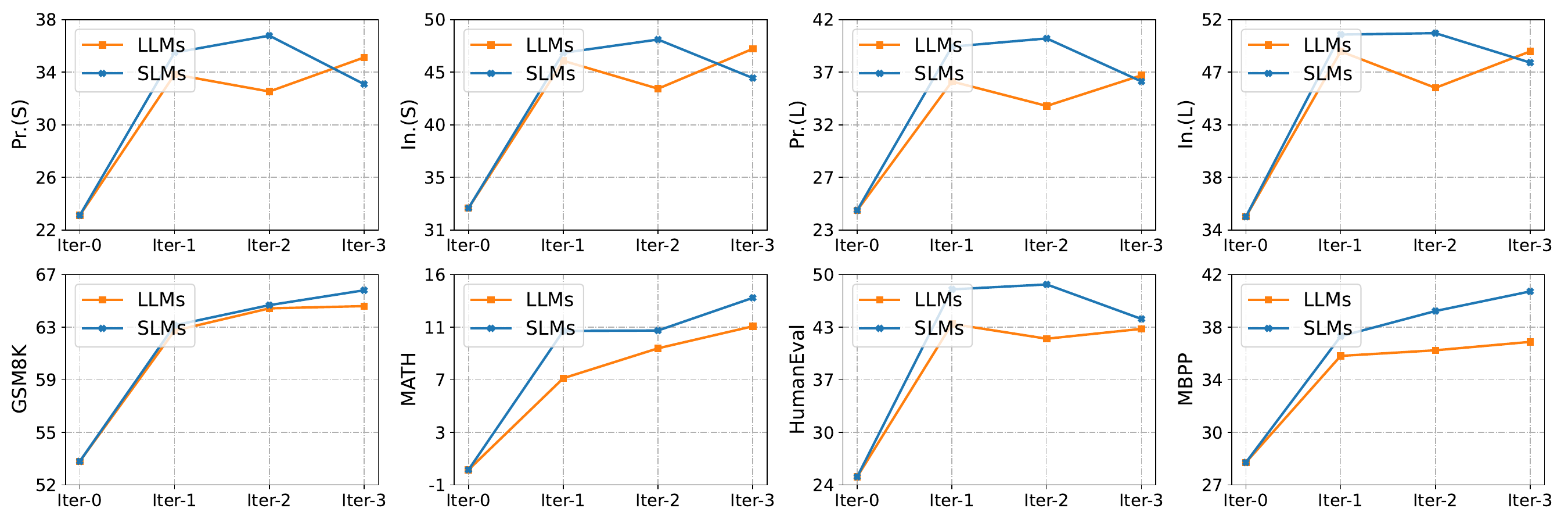}
    \caption{Comparison of performance on Llama-3-8B during three iterations of instruction evolution, using Llama-3.1-8B-Instruct and Llama-3.1-70B-Instruct as supervised models for each round under Evol-Instruct scenario.}  
    \label{fig:iterations}  
\end{figure*}

\begin{figure*}[ht]  
    \centering
    \includegraphics[width=\textwidth]{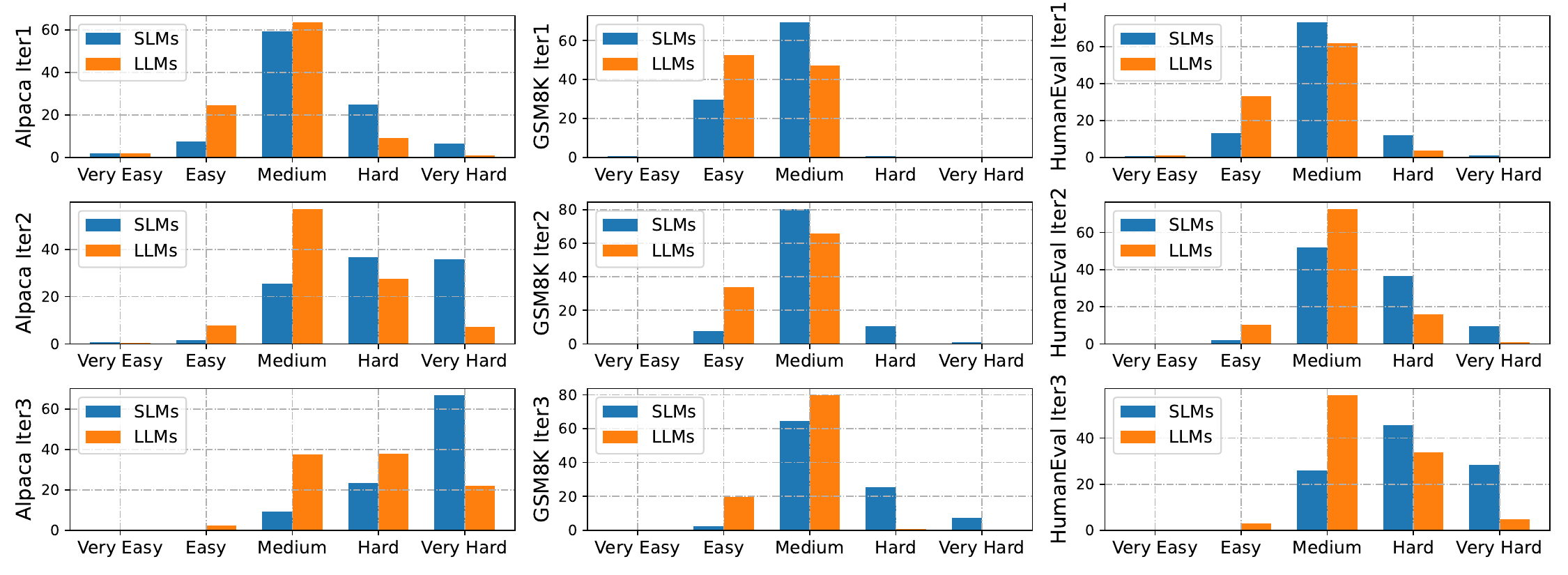}
    \caption{Distribution of difficulty levels for instructions evolved during three iterations, using Llama-3.1-8B-Instruct and Llama-3.1-70B-Instruct as supervised models for each round under Evol-Instruct scenario.}  
    \label{fig:difficulty}  
\end{figure*}

\begin{figure*}[ht]  
    \centering
    \includegraphics[width=\textwidth]{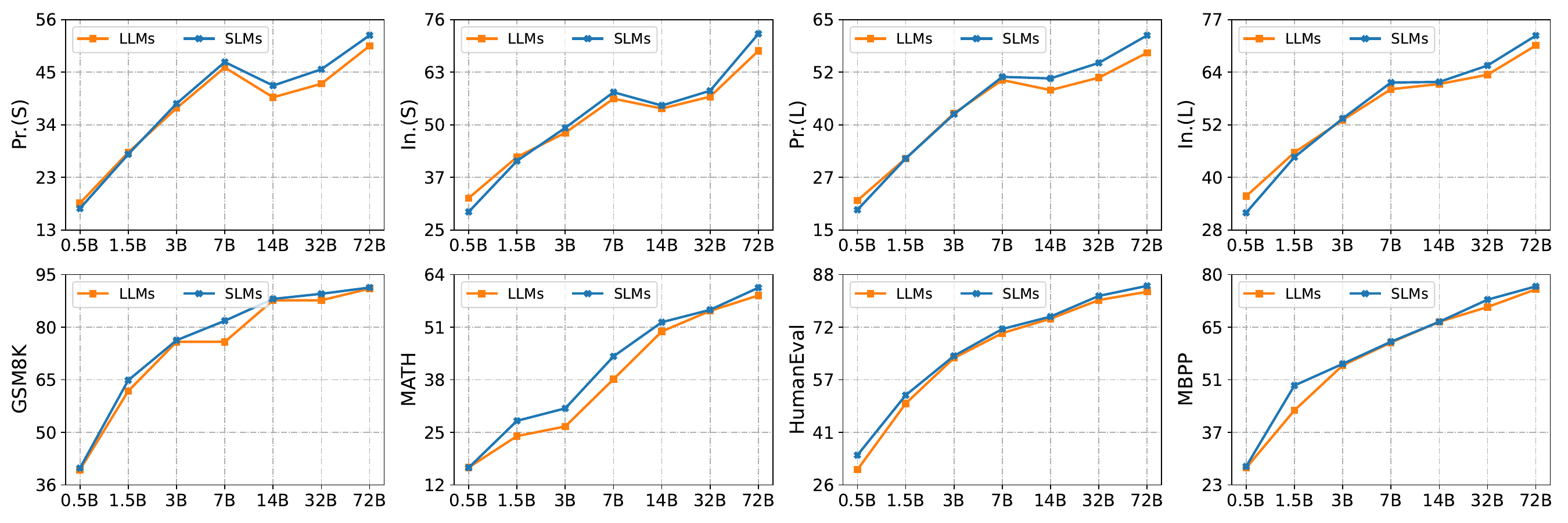}
    \caption{Comparison of performance among Qwen-2.5 series models. Detailed results can be found in Table~\ref{tab:scaling}.}  
    \label{fig:scaling}  
\end{figure*}

\paragraph{Results of Evol-Instruct.} We conduct two sets of experiments, using the Llama-3.1~\citep{DBLP:journals/corr/abs-2407-21783} and Qwen-2~\citep{DBLP:journals/corr/abs-2407-10671} model series for instruction evolution. This approach helps eliminate potential biases specific to each model series, ensuring the generalizability of the conclusions. Specifically, we use Llama-3.1-8B-Instruct and Qwen-2-7B-Instruct as SLMs and Llama-3.1-70B-Instruct and Qwen-2-72B-Instruct serve as LLMs for instruction evolution. To ensure that the generated responses do not influence the experimental conclusions, we consistently use Qwen-2.5-72B-Instruct~\citep{qwen2.5} as the response generator.

Table~\ref{tab:llama_result} and Table~\ref{tab:qwen_result} present a comparative analysis of benchmark results for \textsc{SLM-Inst} and \textsc{LLM-Inst} using the Llama and Qwen model families, highlighting the following key insights\footnote{More results and analyses regarding the performance of seed instruction data and the impact of temperature are provided in Appendix~\ref{sec:more_result}.}.

(1) We find that \textsc{SLM-Inst} outperforms \textsc{LLM-Inst} across instruction following, mathematical reasoning, and code generation, demonstrating superior overall performance in both the Llama and Qwen model families.

(2) More complex and difficult instruction data leads to more effective improvements in instruction following capabilities~\citep{DBLP:journals/corr/abs-2406-13542}. Our results show that \textsc{SLM-Inst} significantly outperforms \textsc{LLM-Inst} on IFEval, highlighting the ability of SLMs to generate more complex instructions compared to LLMs.

\paragraph{Impact of Evolution Iteration.}
Figure~\ref{fig:iterations} illustrates the performance of Llama-3-8B after three rounds of evolution with the Llama-3.1 series (Detailed results can be found in Table~\ref{tab:iterations}). Iter 0 represents the performance of the seed instruction data and we release the following key insights.

(1) We find that during the first two rounds of evolution, the \textsc{SLM-Inst} consistently outperforms \textsc{LLM-Inst}. Notably, in terms of the instruction following, \textsc{LLM-Inst} even experiences negative growth, further proving that SLMs are superior to LLMs in generating complex instructions.

(2) The performance in the third round of evolution shows an interesting phenomenon. While the \textsc{SLM-Inst} continues to perform well in mathematical reasoning, there is a significant drop in both instruction following and code generation. Following~\citep{DBLP:journals/corr/abs-2406-08464}, we use Qwen-2.5-72B-Instruct to assess the difficulty level of the evolved instructions in each round, as shown in Figure~\ref{fig:difficulty}. We find that the difficulty of the \textsc{SLM-Inst} in the third round is excessively high. For example, in the third-round \textsc{SLM-Inst} for Alpaca, nearly 70\% of the instructions are categorized as "very hard". Such overly complex and difficult-to-understand instructions result in a decline in performance. Further data analysis and the evaluation prompt templates can be found in Appendix~\ref{sec:analysis} and Figure~\ref{fig:level}.

(3) We find that the complexity of \textsc{SLM-Inst} in the second iteration surpasses that of \textsc{LLM-Inst} in the third iteration, with \textsc{SLM-Inst} also demonstrating superior performance. This suggests that we can leverage SLMs to generate more complex and challenging instructions with fewer computational resources and evolutionary iterations, while simultaneously achieving better performance.

\paragraph{Scaling Experiments.} To further validate whether our findings hold across models of different sizes, we train models of various sizes within the Qwen-2.5 series (ranging from 0.5B to 72B). The training details can be found in Table~\ref{tab:qwen_hyper}. Due to computational resource constraints, we perform full fine-tuning for models ranging from 0.5B to 7B, while applying LoRA~\citep{DBLP:conf/iclr/HuSWALWWC22} for models from 14B to 72B. To avoid introducing additional biases, we switch the response generator to Llama-3.1-70B-Instruct during the training of the Qwen-2.5 series models. As shown in Figure~\ref{fig:scaling}. We find that in the instruction following evaluation, \textsc{SLM-Inst} performs slightly worse than \textsc{LLM-Inst} on 0.5B and 1.5B models. We believe this is because the evolved instructions in Alpaca are too challenging, and smaller models with lower capabilities may struggle to understand the instructions, leading to performance discrepancies. However, in other evaluations, \textsc{SLM-Inst} shows consistently better performance which further confirms our findings.
\begin{tcolorbox}[colframe=gray!50!black, colback=yellow!10!white, title=\textbf{Finding 1}]
SLMs can evolve more complex and challenging instructions than LLMs.
\end{tcolorbox}

\begin{table*}[ht]
    \centering
    \resizebox{\textwidth}{!}{%
        \begin{tabular}{lcccccccccccccccc}
            \toprule
            \multirow{2}{*}{\textbf{Model}} & \multicolumn{4}{c}{\textbf{IFEval}} & & \multicolumn{6}{c}{\textbf{FollowBench (HSR)}} & & \multicolumn{4}{c}{\textbf{Common Abilities}} \\
            \cmidrule{2-17}
            & \textbf{Pr.(S)} & \textbf{In.(S)} & \textbf{Pr.(L)} & \textbf{In.(L)} & & \textbf{Level 1} & \textbf{Level 2} & \textbf{Level 3} & \textbf{Level 4} & \textbf{Level 5} & \textbf{Avg.} & & \textbf{C-Eval} & \textbf{MMLU} & \textbf{HumanEval} & \textbf{GSM8K} \\
            \midrule
            \multicolumn{17}{c}{\textit{Supervision Model: Llama-3.1-70B-Instruct}} \\
            Llama-3.2-3B & 40.85 & 51.92 & 42.33 & 53.84 & \vrule & \textbf{61.17} & 57.59 & \textbf{50.55} & 33.09 & 26.74 & 45.83 & \vrule & \textbf{41.37} & \textbf{52.65} & \textbf{29.88} & 27.07 \\
            Llama-3-8B & 37.71 & 50.00 & 39.19 & 52.04 & \vrule & 49.64 & 46.60 & 41.56 & 27.05 & 22.37 & 37.44 & \vrule & 41.87 & 51.14 & 26.83 & 37.76 \\
            Llama-3.1-8B & 41.96 & 53.36 & 42.70 & 54.20 & \vrule & 51.77 & 45.60 & 45.04 & 34.85 & 26.61 & 40.78 & \vrule & \textbf{44.50} & 56.39 & 31.10 & 38.21 \\
            Qwen-2-7B & 41.96 & 53.60 & 43.62 & 55.64 & \vrule & 72.18 & 62.45 & \textbf{56.43} & 41.31 & 35.42 & 53.56 & \vrule & \textbf{81.08} & 55.71 & 57.32 & \textbf{79.68} \\
            Qwen-2.5-7B & 49.17 & \textbf{60.31} & 50.46 & 61.51 & \vrule & \textbf{78.88} & \textbf{73.78} & \textbf{61.50} & 51.99 & 45.42 & \textbf{62.31} & \vrule & \textbf{80.46} & 58.39 & 67.68 & \textbf{85.90} \\
            InternLM-2-7B & 46.21 & 56.71 & 48.06 & 58.63 & \vrule & 68.89 & 62.23 & 54.17 & 44.27 & 42.06 & 54.33 & \vrule & 60.11 & 60.59 & 65.35 & 50.00 \\
            \midrule
            \multicolumn{17}{c}{\textit{Supervision Model: Llama-3.1-8B-Instruct}} \\
            Llama-3.2-3B & \textbf{43.62} & \textbf{54.20} & \textbf{46.95} & \textbf{57.07} & \vrule & 56.95 & \textbf{61.46} & 50.20 & \textbf{37.65} & \textbf{34.16} & \textbf{48.08} & \vrule & 40.56 & 49.08 & 25.00 & \textbf{29.87} \\
            Llama-3-8B & \textbf{41.04} & \textbf{51.32} & \textbf{42.88} & \textbf{53.11} & \vrule & \textbf{62.99} & \textbf{54.38} & \textbf{49.29} & \textbf{32.21} & \textbf{32.21} & \textbf{46.21} & \vrule & \textbf{43.49} & \textbf{55.63} & \textbf{37.20} & \textbf{45.26} \\
            Llama-3.1-8B & \textbf{42.51} & \textbf{54.92} & \textbf{44.73} & \textbf{56.71} & \vrule & \textbf{63.99} & \textbf{58.15} & \textbf{53.29} & \textbf{39.49} & \textbf{36.02} & \textbf{50.19} & \vrule & 43.77 & \textbf{58.32} & \textbf{32.32} & \textbf{47.92} \\
            Qwen-2-7B & \textbf{44.92} & \textbf{55.76} & \textbf{47.50} & \textbf{58.39} & \vrule & \textbf{78.75} & \textbf{63.30} & 52.31 & \textbf{50.28} & \textbf{43.08} & \textbf{57.54} & \vrule & 80.11 & \textbf{56.84} & \textbf{65.24} & 79.53 \\
            Qwen-2.5-7B & \textbf{50.09} & 59.59 & \textbf{52.50} & \textbf{61.75} & \vrule & 77.86 & 70.22 & 59.86 & \textbf{53.35} & \textbf{47.18} & 61.69 & \vrule & 79.74 & \textbf{60.17} & \textbf{72.56} & 84.69 \\
            InternLM-2-7B & \textbf{47.50} & \textbf{57.67} & \textbf{50.83} & \textbf{61.15} & \vrule & \textbf{74.73} & \textbf{66.16} & \textbf{61.94} & \textbf{54.10} & \textbf{46.28} & \textbf{60.64} & \vrule & \textbf{63.03} & \textbf{63.16} & \textbf{70.96} & \textbf{54.27} \\
            \bottomrule
        \end{tabular}
    }
    \caption{Comparison of performance with Llama-3.1-8B-Instruct and Llama-3.1-70B-Instruct as supervised models under AutoIF scenario.}
    \label{tab:autoif}
\end{table*}
\subsection{AutoIF Scenario}
In this section, we mainly concentrate on \textit{whether SLMs can generate more diverse instruction data compared to LLMs.}

\paragraph{Evaluation Benchmarks and Metrics.} We fully adhere to the evaluation benchmarks used in AutoIF. Specifically, we utilize IFEval and FollowBench~\citep{DBLP:conf/acl/Jiang0ZZLMS00W24} to assess instruction following capabilities\footnote{We use the Microsoft Azure OpenAI GPT-4 API.}. We also evaluate our models on C-Eval~\citep{DBLP:conf/nips/HuangBZZZSLLZLF23}, MMLU~\citep{DBLP:conf/iclr/HendrycksBBZMSS21}, GSM8K, and HumanEval to obtain a comprehensive assessment of their capabilities. For detailed information, please refer to Appendix~\ref{sec:evaluations}.

\paragraph{Results of AutoIF.} We use the Llama-3.1 series models for synthesizing instructions and we adopt Qwen-2.5-72B-Instruct for generating responses under the AutoIF scenario. As shown in Table~\ref{tab:autoif}, on the IFEval and FollowBench instruction following benchmarks, the instruction data augmented by SLMs achieved better performance. Especially on FollowBench, \textsc{SLM-Inst} even achieve nearly a 10\% improvement over Llama-3-8B and Llama-3.1-8B. Meanwhile, on common abilities, \textsc{SLM-Inst} also demonstrates competitive performance.

\begin{figure}[ht]  
    \centering
    \includegraphics[width=0.48\textwidth]{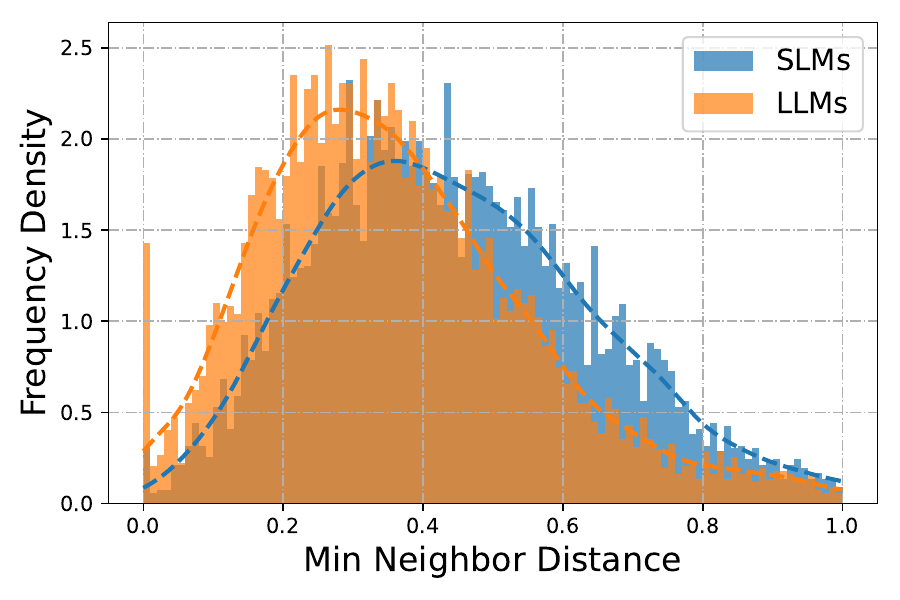}
    \caption{Distribution of Minimum Neighbor Distance for instructions generated by Llama-3.1-8B-Instruct and Llama-3.1-70B-Instruct in the AutoIF scenario.}  
    \label{fig:diversity}  
\end{figure}

AutoIF begins with a small set of manually crafted seed instructions, from which the model draws inspiration to generate a large number of new instructions and perform verifications to ensure their quality. Since the generated instructions have undergone multiple rounds of verification, their diversity becomes even more crucial. Following~\citep{DBLP:journals/corr/abs-2406-08464}, we use all-mpnet-base-v2~\citep{DBLP:conf/nips/Song0QLL20} to measure similarity via minimum neighbor distance (MND) in the embedding space. Notably, a high number of samples with low MND suggests poor diversity within the dataset. Figure~\ref{fig:diversity} demonstrates that \textsc{SLM-Inst} has more samples with a larger MND, indicating higher diversity than \textsc{LLM-Inst}.
\begin{tcolorbox}[colframe=gray!50!black, colback=yellow!10!white, title=\textbf{Finding 2}]
SLMs can generate more diverse instructions than LLMs.
\end{tcolorbox}

\begin{table*}[ht]
    \centering
    \small
        \begin{tabular}{lcccccccccc}
            \toprule
            \multirow{2}{*}{\textbf{Model}} & \multicolumn{4}{c}{\textbf{Instruction Following (IFEval)}} & & \multicolumn{2}{c}{\textbf{Math Reasoning}} & & \multicolumn{2}{c}{\textbf{Code Generation}} \\
            \cmidrule{2-11}
            & \textbf{Pr.(S)} & \textbf{In.(S)} & \textbf{Pr.(L)} & \textbf{In.(L)} & & \textbf{GSM8K} & \textbf{MATH} & & \textbf{HumanEval} & \textbf{MBPP} \\
            \midrule
            \multicolumn{11}{c}{\textit{Supervised Model: Llama-3.1-70B-Instruct}}  \\
            Llama-3.2-3B & 36.60 & 48.68 & 39.00 & 51.08 & \vrule & 53.60 & 7.56 & \vrule & 35.37 & 33.00 \\
            Llama-3-8B & 35.86 & 47.60 & 38.63 & 50.24 & \vrule & 63.91 & 9.18 & \vrule & 38.41 & 32.40 \\
            Llama-3.1-8B & 36.97 & 47.60 & 40.30 & 51.08 & \vrule & 66.11 & 11.68 & \vrule & 40.85 & \textbf{40.40} \\
            \midrule
            \multicolumn{11}{c}{\textit{Supervised Model: Llama-3.1-8B-Instruct}} \\
            Llama-3.2-3B & \textbf{45.47} & \textbf{57.43} & \textbf{50.28} & \textbf{61.27} & \vrule & \textbf{56.48} & \textbf{8.42} & \vrule & \textbf{38.41} & \textbf{34.40} \\
            Llama-3-8B & \textbf{37.34} & \textbf{49.64} & \textbf{39.74} & \textbf{51.56} & \vrule & \textbf{67.40} & \textbf{12.26} & \vrule & \textbf{43.90} & \textbf{34.80} \\
            Llama-3.1-8B & \textbf{38.08} & \textbf{49.76} & \textbf{40.48} & \textbf{52.40} & \vrule & \textbf{69.52} & \textbf{15.62} & \vrule & \textbf{51.22} & 38.80 \\
            \bottomrule
        \end{tabular}
    \caption{Comparison of performance with Llama-3.1-8B-Instruct and Llama-3.1-70B-Instruct as supervised models under Auto Evol-Instruct scenario.}
    \label{tab:auto_evol}
\end{table*}
\subsection{Auto Evol-Instruct Scenario}
In this section, we mainly focus on \textit{whether SLMs can automatically evolve more effective instructions compared to LLMs.} 

\paragraph{Results of Auto Evol-Instruct.} As shown in Table~\ref{tab:auto_evol}, we find that the instruction data automatically evolved by SLMs consistently performs better across the Llama series models than LLMs. In addition, we prompt the Qwen-2.5-72B-Instruct model to summarize and deduplicate keywords from the trajectories generated by SLMs and LLMs (the prompt template can be found in Figure~\ref{fig:extract}). We find that the number of trajectories produced by SLMs is 6.9\% higher than that of LLMs, further highlighting that SLMs can design more varied evolutionary trajectories, leading to more complex and diverse instructions.
\begin{tcolorbox}[colframe=gray!50!black, colback=yellow!10!white, title=\textbf{Finding 3}]
SLMs can automatically evolve more effective instructions than LLMs.
\end{tcolorbox}

\begin{figure}[ht]  
    \centering
    \includegraphics[width=0.48\textwidth]{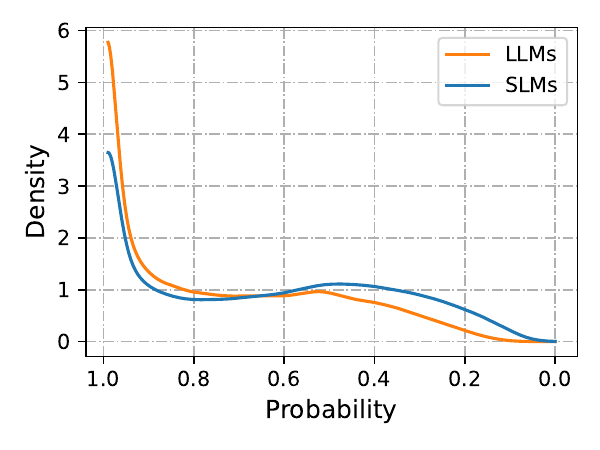}
    \caption{Comparison of output token probability distributions in the Evol-Instruct scenario.}  
    \label{fig:logprob}  
\end{figure}

\section{RQ2: Why Do SLMs Outperform LLMs in Evolving Instructions?}
In this section, we primarily analyze why SLMs perform better from the perspectives of model inference and real-world cases.

\paragraph{Comparison of Token Distributions.} The results of our previous experiments indicate that SLMs are capable of evolving and generating more complex and diverse instructions. We hypothesize that this is due to the superior instruction following capabilities of LLMs, which result in a narrower output space (overconfidence) when following instructions, thereby leading to less diversity and complexity in the generated new instructions. To validate this hypothesis, we employ the Llama-3.1-8B-Instruct and Llama-3.1-70B-Instruct models within the Evol-Instruct scenario to obtain the probability distributions of output tokens. By extracting the top-1 token probability at each output position, we compare the output probability distributions between SLMs and LLMs. As shown in Figure~\ref{fig:logprob}, we observe that the top-1 token output probability for SLMs is lower, suggesting that the output distribution of SLMs is more diverse. This supports our hypothesis that, due to their relatively weaker instruction following capabilities compared to LLMs, SLMs generate a broader output space, leading to more diverse and complex instructions. We also analyze some cases, and the detailed results can be found in Appendix~\ref{sec:more_result}.
\begin{tcolorbox}[colframe=gray!50!black, colback=yellow!10!white, title=\textbf{Finding 4}]
SLMs have a broader output space and are less likely to be overconfident than LLMs.
\end{tcolorbox}

\section{RQ3: How Do We Determine Whether An Instruction is Effective without Instruction Tuning?}
In this section, we primarily discuss how to determine whether instruction data is effective without instruction tuning. 

\paragraph{Instruction Complex-Aware IFD.} As mentioned in~\citep{xu2024strongermodelsstrongerteachers}, existing evaluations typically focus on assessing responses, such as using reward models, while neglecting the impact of instructions on the data. Recently, \citet{DBLP:conf/naacl/LiZLCC0W0024} proposed the instruction following Difficulty (IFD) score to evaluate the quality of instructions. Specifically, the formula for IFD is as follows.
\begin{equation}
    \text{IFD}_{\Theta}(Q, A) = \frac{L_{\Theta}(A | Q)}{L_{\Theta}(A)}
\end{equation}
Where $Q$ and $A$ represent instructions and responses, and $L_{\Theta}(\cdot)$ represents the average cross entropy loss determined by a model $\Theta$. IFD can be understood as the importance of instructions in generating responses. A lower IFD means that a sample does not require training, as the model is already able to generate the corresponding response effectively when given the instruction. However, as shown in Figure~\ref{fig:iterations} and Table~\ref{tab:ifd}, when the difficulty of the instructions is too high, it may result in a higher IFD, but the overall performance may fall short of expectations. Inspired by this, we introduce the difficulty level of instructions into the original IFD and propose the Instruction Complex-Aware IFD (IC-IFD). Specifically, we introduce the perplexity of the instructions into the original IFD score, resulting in the following formula.
\begin{equation}
    \text{IC-IFD}_{\Theta}(Q, A) = \frac{L_{\Theta}(A | Q)}{L_{\Theta}(Q) \cdot L_{\Theta}(A)} 
\end{equation}

\begin{table}[ht]
\centering
\resizebox{0.48\textwidth}{!}{%
\begin{tabular}{lcccc}
\toprule
\multirow{2}{*}{\textbf{Metrics}} & \multicolumn{4}{c}{\textbf{IFEval}}                                   \\ \cmidrule{2-5} 
                                  & \textbf{Pr.(S)} & \textbf{In.(S)} & \textbf{Pr.(L)} & \textbf{In.(L)} \\ \midrule
Original                          & 33.09           & 44.72           & 36.41           & 48.32           \\
Instruction Len.                  & 29.94       & 39.69                & 33.83                & 43.53                \\
Instruction PPL                   & 27.91                & 39.69       & 32.35       & 44.36                \\
IFD                               & 30.87                & 43.53                & 36.04                & 47.60                \\
IC-IFD                            & \textbf{34.01}  & \textbf{46.16}  & \textbf{38.82}  & \textbf{50.72}  \\ \bottomrule
\end{tabular}
}
\caption{Comparison of different metrics under 25\% of Alpaca-iter3 evolved by SLMs on Llama-3-8B.}
\label{tab:icifd}
\end{table}

\paragraph{Performance of IC-IFD.} To validate the effectiveness of IC-IFD, we aim to mitigate the performance degradation caused by the third round of instruction data evolved by SLMs. Specifically, we retain the top 25\% of instruction data using several metrics, including instruction length (filtering out overly long instructions), instruction perplexity (PPL, filtering out instructions with excessively high PPL), IFD, and IC-IFD. As shown in Table~\ref{tab:icifd}, under the condition of retaining only 25\% of the instruction data, IC-IFD outperforms the full dataset, while other metrics exhibit varying degrees of performance degradation, thereby demonstrating the effectiveness of IC-IFD. Further experiments on IC-IFD can be found in Appendix~\ref{sec:more_result}.

\section{Related Work}
Instruction tuning has become an essential strategy for enhancing the capabilities of large language models (LLMs)~\citep{DBLP:conf/nips/Ouyang0JAWMZASR22, DBLP:journals/corr/abs-2303-08774}. By curating high-quality datasets, we can more effectively align these models with specific objectives~\citep{DBLP:conf/nips/ZhouLX0SMMEYYZG23}. Recently, some researchers have highlighted the significance of instruction data that is either manually annotated or developed with human involvement, such as ShareGPT~\citep{vicuna2023} and OpenAssistant~\citep{DBLP:conf/nips/KopfKRATSBNSNES23}. Meanwhile, other studies concentrate on leveraging LLMs to generate high-quality datasets with minimal human effort~\citep{DBLP:conf/iclr/XuSZG0FTLJ24,DBLP:conf/iclr/LuoX0SGHT0LJ24,DBLP:journals/corr/abs-2308-09583}. \citet{DBLP:conf/acl/WangKMLSKH23} introduces Self-Instruct, which begins with a small collection of manually crafted seed instructions and utilizes LLMs to expand these instructions, ultimately producing a large-scale instruction set that improves model abilities. \citet{DBLP:conf/iclr/XuSZG0FTLJ24} presents Evol-Instruct, which employs LLMs for the iterative enhancement of the original instructions through both in-depth and breadth evolution, resulting in a more complex and diverse instruction dataset. Auto Evol-Instruct~\citep{DBLP:conf/emnlp/ZengXZLC24} further removes human involvement, enabling LLMs to autonomously design the evolution trajectory based on the original instructions. AutoIF~\citep{DBLP:journals/corr/abs-2406-13542} introduces a code feedback mechanism that allows LLMs to generate evaluation code for verifying whether the quality of the instructions meets the required standards. \citet{DBLP:journals/corr/abs-2406-08464} only provides a single prompt to induce the model to generate a large amount of instruction data. Current research primarily focuses on utilizing larger language models, such as GPT-4~\citep{DBLP:journals/corr/abs-2303-08774}, for constructing complex instructions. More recently, \citet{xu2024strongermodelsstrongerteachers} explores the performance differences of various-sized models as response generators. In contrast, we concentrate on the potential of smaller language models in evolving complex instructions. This innovation not only reduces the costs associated with instructions construction but, more importantly, offers a comprehensive evaluation and exploration, highlighting the significant capabilities inherent in smaller models and providing valuable insights for future work.

\section{Conclusion}
In this paper, we compare the performance of SLMs and LLMs in evolving instructions. Extensive experiments demonstrate that SLMs can synthesize more effective instructions at a lower computational cost than LLMs. Through an analysis of the model output distributions, we observe that SLMs exhibit a broader output space, leading to more complex and diverse instructions. Furthermore, we introduce instruction complexity as a penalty term in the original IFD and propose IC-IFD, which allows for more accurate assessment of instruction data effectiveness without the need for instruction tuning. Our work also lays the groundwork for future research on SLMs in instruction data synthesis, offering a foundation understanding for further exploration.

\newpage
\section*{Limitations}
Although our work provides valuable insights that SLMs perform better in evolving instructions through comprehensive experiments, several directions are worth exploring in future research.

(1) We have only conducted experiments in instruction following, mathematical reasoning, and code generation. We have not focused on other broader domains, and there may have interesting discoveries in these areas that require future work.

(2) Our work focuses on comparing SLMs and LLMs in evolving instruction sets, rather than exploring the full potential of SLMs in synthesizing entire instruction datasets. Future research that investigates the capabilities of SLMs across the entire instruction data synthesis pipeline would be a promising and exciting direction to explore.

(3) The IC-IFD we propose is based on our observation that performance degrades with the emergence of high-difficulty instructions, which leads us to introduce instruction complexity as a penalty term in the original IFD. In the future, further exploration into how to more accurately assess the effectiveness of instruction data without instruction tuning would be valuable.

\section*{Acknowledgments}
This research is supported by the National Natural Science Foundation of China (62072052) and the Innovation Research Group Project of NSFC (61921003). At the same time, we sincerely appreciate the academic and computational support from the Beijing Academy of Artificial Intelligence (BAAI), which has been crucial to the successful completion of this study.

\bibliography{custom}

\begin{thebibliography}{40}
\providecommand{\natexlab}[1]{#1}

\bibitem[{Austin et~al.(2021)Austin, Odena, Nye, Bosma, Michalewski, Dohan, Jiang, Cai, Terry, Le, and Sutton}]{DBLP:journals/corr/abs-2108-07732}
Jacob Austin, Augustus Odena, Maxwell~I. Nye, Maarten Bosma, Henryk Michalewski, David Dohan, Ellen Jiang, Carrie~J. Cai, Michael Terry, Quoc~V. Le, and Charles Sutton. 2021.
\newblock \href {https://arxiv.org/abs/2108.07732} {Program synthesis with large language models}.
\newblock \emph{CoRR}, abs/2108.07732.

\bibitem[{Bi et~al.(2024)Bi, Chen, Chen, Chen, Dai, Deng, Ding, Dong, Du, Fu, Gao, Gao, Gao, Ge, Guan, Guo, Guo, Hao, Hao, He, Hu, Huang, Li, Li, Li, Li, Li, Liang, Lin, Liu, Liu, Liu, Liu, Liu, Liu, Lu, Lu, Luo, Ma, Nie, Pei, Piao, Qiu, Qu, Ren, Ren, Ruan, Sha, Shao, Song, Su, Sun, Sun, Tang, Wang, Wang, Wang, Wang, Wang, Wu, Wu, Xie, Xie, Xie, Xiong, Xu, Xu, Xu, Yang, You, Yu, Yu, Zhang, Zhang, Zhang, Zhang, Zhang, Zhang, Zhang, Zhang, Zhao, Zhao, Zhou, Zhou, Zhu, and Zou}]{DBLP:journals/corr/abs-2401-02954}
Xiao Bi, Deli Chen, Guanting Chen, Shanhuang Chen, Damai Dai, Chengqi Deng, Honghui Ding, Kai Dong, Qiushi Du, Zhe Fu, Huazuo Gao, Kaige Gao, Wenjun Gao, Ruiqi Ge, Kang Guan, Daya Guo, Jianzhong Guo, Guangbo Hao, Zhewen Hao, Ying He, Wenjie Hu, Panpan Huang, Erhang Li, Guowei Li, Jiashi Li, Yao Li, Y.~K. Li, Wenfeng Liang, Fangyun Lin, Alex~X. Liu, Bo~Liu, Wen Liu, Xiaodong Liu, Xin Liu, Yiyuan Liu, Haoyu Lu, Shanghao Lu, Fuli Luo, Shirong Ma, Xiaotao Nie, Tian Pei, Yishi Piao, Junjie Qiu, Hui Qu, Tongzheng Ren, Zehui Ren, Chong Ruan, Zhangli Sha, Zhihong Shao, Junxiao Song, Xuecheng Su, Jingxiang Sun, Yaofeng Sun, Minghui Tang, Bingxuan Wang, Peiyi Wang, Shiyu Wang, Yaohui Wang, Yongji Wang, Tong Wu, Y.~Wu, Xin Xie, Zhenda Xie, Ziwei Xie, Yiliang Xiong, Hanwei Xu, R.~X. Xu, Yanhong Xu, Dejian Yang, Yuxiang You, Shuiping Yu, Xingkai Yu, B.~Zhang, Haowei Zhang, Lecong Zhang, Liyue Zhang, Mingchuan Zhang, Minghua Zhang, Wentao Zhang, Yichao Zhang, Chenggang Zhao, Yao Zhao, Shangyan Zhou, Shunfeng Zhou, Qihao
  Zhu, and Yuheng Zou. 2024.
\newblock \href {https://doi.org/10.48550/ARXIV.2401.02954} {Deepseek {LLM:} scaling open-source language models with longtermism}.
\newblock \emph{CoRR}, abs/2401.02954.

\bibitem[{Cai et~al.(2024)Cai, Cao, Chen, Chen, Chen, Chen, Chen, Chen, Chen, Chu, Dong, Duan, Fan, Fei, Gao, Ge, Gu, Gu, Gui, Guo, Guo, He, Hu, Huang, Jiang, Jiao, Jin, Lei, Li, Li, Li, Li, Li, Li, Liu, Liu, Hong, Liu, Liu, Liu, Lv, Lv, Lv, Ma, Ma, Ma, Ning, Ouyang, Qiu, Qu, Shang, Shao, Song, Song, Sui, Sun, Sun, Tang, Wang, Wang, Wang, Wang, Wang, Wang, Wang, Wei, Weng, Wu, Xiong, Zhao, and et~al.}]{DBLP:journals/corr/abs-2403-17297}
Zheng Cai, Maosong Cao, Haojiong Chen, Kai Chen, Keyu Chen, Xin Chen, Xun Chen, Zehui Chen, Zhi Chen, Pei Chu, Xiaoyi Dong, Haodong Duan, Qi~Fan, Zhaoye Fei, Yang Gao, Jiaye Ge, Chenya Gu, Yuzhe Gu, Tao Gui, Aijia Guo, Qipeng Guo, Conghui He, Yingfan Hu, Ting Huang, Tao Jiang, Penglong Jiao, Zhenjiang Jin, Zhikai Lei, Jiaxing Li, Jingwen Li, Linyang Li, Shuaibin Li, Wei Li, Yining Li, Hongwei Liu, Jiangning Liu, Jiawei Hong, Kaiwen Liu, Kuikun Liu, Xiaoran Liu, Chengqi Lv, Haijun Lv, Kai Lv, Li~Ma, Runyuan Ma, Zerun Ma, Wenchang Ning, Linke Ouyang, Jiantao Qiu, Yuan Qu, Fukai Shang, Yunfan Shao, Demin Song, Zifan Song, Zhihao Sui, Peng Sun, Yu~Sun, Huanze Tang, Bin Wang, Guoteng Wang, Jiaqi Wang, Jiayu Wang, Rui Wang, Yudong Wang, Ziyi Wang, Xingjian Wei, Qizhen Weng, Fan Wu, Yingtong Xiong, Xiaomeng Zhao, and et~al. 2024.
\newblock \href {https://doi.org/10.48550/ARXIV.2403.17297} {Internlm2 technical report}.
\newblock \emph{CoRR}, abs/2403.17297.

\bibitem[{Chaudhary(2023)}]{codealpaca}
Sahil Chaudhary. 2023.
\newblock Code alpaca: An instruction-following llama model for code generation.
\newblock \url{https://github.com/sahil280114/codealpaca}.

\bibitem[{Chen et~al.(2021)Chen, Tworek, Jun, Yuan, de~Oliveira~Pinto, Kaplan, Edwards, Burda, Joseph, Brockman, Ray, Puri, Krueger, Petrov, Khlaaf, Sastry, Mishkin, Chan, Gray, Ryder, Pavlov, Power, Kaiser, Bavarian, Winter, Tillet, Such, Cummings, Plappert, Chantzis, Barnes, Herbert{-}Voss, Guss, Nichol, Paino, Tezak, Tang, Babuschkin, Balaji, Jain, Saunders, Hesse, Carr, Leike, Achiam, Misra, Morikawa, Radford, Knight, Brundage, Murati, Mayer, Welinder, McGrew, Amodei, McCandlish, Sutskever, and Zaremba}]{DBLP:journals/corr/abs-2107-03374}
Mark Chen, Jerry Tworek, Heewoo Jun, Qiming Yuan, Henrique~Pond{\'{e}} de~Oliveira~Pinto, Jared Kaplan, Harri Edwards, Yuri Burda, Nicholas Joseph, Greg Brockman, Alex Ray, Raul Puri, Gretchen Krueger, Michael Petrov, Heidy Khlaaf, Girish Sastry, Pamela Mishkin, Brooke Chan, Scott Gray, Nick Ryder, Mikhail Pavlov, Alethea Power, Lukasz Kaiser, Mohammad Bavarian, Clemens Winter, Philippe Tillet, Felipe~Petroski Such, Dave Cummings, Matthias Plappert, Fotios Chantzis, Elizabeth Barnes, Ariel Herbert{-}Voss, William~Hebgen Guss, Alex Nichol, Alex Paino, Nikolas Tezak, Jie Tang, Igor Babuschkin, Suchir Balaji, Shantanu Jain, William Saunders, Christopher Hesse, Andrew~N. Carr, Jan Leike, Joshua Achiam, Vedant Misra, Evan Morikawa, Alec Radford, Matthew Knight, Miles Brundage, Mira Murati, Katie Mayer, Peter Welinder, Bob McGrew, Dario Amodei, Sam McCandlish, Ilya Sutskever, and Wojciech Zaremba. 2021.
\newblock \href {https://arxiv.org/abs/2107.03374} {Evaluating large language models trained on code}.
\newblock \emph{CoRR}, abs/2107.03374.

\bibitem[{Chiang et~al.(2023)Chiang, Li, Lin, Sheng, Wu, Zhang, Zheng, Zhuang, Zhuang, Gonzalez, Stoica, and Xing}]{vicuna2023}
Wei-Lin Chiang, Zhuohan Li, Zi~Lin, Ying Sheng, Zhanghao Wu, Hao Zhang, Lianmin Zheng, Siyuan Zhuang, Yonghao Zhuang, Joseph~E. Gonzalez, Ion Stoica, and Eric~P. Xing. 2023.
\newblock \href {https://lmsys.org/blog/2023-03-30-vicuna/} {Vicuna: An open-source chatbot impressing gpt-4 with 90\%* chatgpt quality}.

\bibitem[{Cobbe et~al.(2021)Cobbe, Kosaraju, Bavarian, Chen, Jun, Kaiser, Plappert, Tworek, Hilton, Nakano, Hesse, and Schulman}]{DBLP:journals/corr/abs-2110-14168}
Karl Cobbe, Vineet Kosaraju, Mohammad Bavarian, Mark Chen, Heewoo Jun, Lukasz Kaiser, Matthias Plappert, Jerry Tworek, Jacob Hilton, Reiichiro Nakano, Christopher Hesse, and John Schulman. 2021.
\newblock \href {https://arxiv.org/abs/2110.14168} {Training verifiers to solve math word problems}.
\newblock \emph{CoRR}, abs/2110.14168.

\bibitem[{Contributors(2023)}]{2023opencompass}
OpenCompass Contributors. 2023.
\newblock Opencompass: A universal evaluation platform for foundation models.
\newblock \url{https://github.com/open-compass/opencompass}.

\bibitem[{Dong et~al.(2024)Dong, Lu, Li, Xia, Yu, Zhou, and Zhou}]{DBLP:journals/corr/abs-2406-13542}
Guanting Dong, Keming Lu, Chengpeng Li, Tingyu Xia, Bowen Yu, Chang Zhou, and Jingren Zhou. 2024.
\newblock \href {https://doi.org/10.48550/ARXIV.2406.13542} {Self-play with execution feedback: Improving instruction-following capabilities of large language models}.
\newblock \emph{CoRR}, abs/2406.13542.

\bibitem[{Dubey et~al.(2024)Dubey, Jauhri, Pandey, Kadian, Al{-}Dahle, Letman, Mathur, Schelten, Yang, Fan, Goyal, Hartshorn, Yang, Mitra, Sravankumar, Korenev, Hinsvark, Rao, Zhang, Rodriguez, Gregerson, Spataru, Rozi{\`{e}}re, Biron, Tang, Chern, Caucheteux, Nayak, Bi, Marra, McConnell, Keller, Touret, Wu, Wong, Ferrer, Nikolaidis, Allonsius, Song, Pintz, Livshits, Esiobu, Choudhary, Mahajan, Garcia{-}Olano, Perino, Hupkes, Lakomkin, AlBadawy, Lobanova, Dinan, Smith, Radenovic, Zhang, Synnaeve, Lee, Anderson, Nail, Mialon, Pang, Cucurell, Nguyen, Korevaar, Xu, Touvron, Zarov, Ibarra, Kloumann, Misra, Evtimov, Copet, Lee, Geffert, Vranes, Park, Mahadeokar, Shah, van~der Linde, Billock, Hong, Lee, Fu, Chi, Huang, Liu, Wang, Yu, Bitton, Spisak, Park, Rocca, Johnstun, Saxe, Jia, Alwala, Upasani, Plawiak, Li, Heafield, Stone, and et~al.}]{DBLP:journals/corr/abs-2407-21783}
Abhimanyu Dubey, Abhinav Jauhri, Abhinav Pandey, Abhishek Kadian, Ahmad Al{-}Dahle, Aiesha Letman, Akhil Mathur, Alan Schelten, Amy Yang, Angela Fan, Anirudh Goyal, Anthony Hartshorn, Aobo Yang, Archi Mitra, Archie Sravankumar, Artem Korenev, Arthur Hinsvark, Arun Rao, Aston Zhang, Aur{\'{e}}lien Rodriguez, Austen Gregerson, Ava Spataru, Baptiste Rozi{\`{e}}re, Bethany Biron, Binh Tang, Bobbie Chern, Charlotte Caucheteux, Chaya Nayak, Chloe Bi, Chris Marra, Chris McConnell, Christian Keller, Christophe Touret, Chunyang Wu, Corinne Wong, Cristian~Canton Ferrer, Cyrus Nikolaidis, Damien Allonsius, Daniel Song, Danielle Pintz, Danny Livshits, David Esiobu, Dhruv Choudhary, Dhruv Mahajan, Diego Garcia{-}Olano, Diego Perino, Dieuwke Hupkes, Egor Lakomkin, Ehab AlBadawy, Elina Lobanova, Emily Dinan, Eric~Michael Smith, Filip Radenovic, Frank Zhang, Gabriel Synnaeve, Gabrielle Lee, Georgia~Lewis Anderson, Graeme Nail, Gr{\'{e}}goire Mialon, Guan Pang, Guillem Cucurell, Hailey Nguyen, Hannah Korevaar, Hu~Xu, Hugo
  Touvron, Iliyan Zarov, Imanol~Arrieta Ibarra, Isabel~M. Kloumann, Ishan Misra, Ivan Evtimov, Jade Copet, Jaewon Lee, Jan Geffert, Jana Vranes, Jason Park, Jay Mahadeokar, Jeet Shah, Jelmer van~der Linde, Jennifer Billock, Jenny Hong, Jenya Lee, Jeremy Fu, Jianfeng Chi, Jianyu Huang, Jiawen Liu, Jie Wang, Jiecao Yu, Joanna Bitton, Joe Spisak, Jongsoo Park, Joseph Rocca, Joshua Johnstun, Joshua Saxe, Junteng Jia, Kalyan~Vasuden Alwala, Kartikeya Upasani, Kate Plawiak, Ke~Li, Kenneth Heafield, Kevin Stone, and et~al. 2024.
\newblock \href {https://doi.org/10.48550/ARXIV.2407.21783} {The llama 3 herd of models}.
\newblock \emph{CoRR}, abs/2407.21783.

\bibitem[{Dubois et~al.(2023)Dubois, Li, Taori, Zhang, Gulrajani, Ba, Guestrin, Liang, and Hashimoto}]{DBLP:conf/nips/DuboisLTZGBGLH23}
Yann Dubois, Chen~Xuechen Li, Rohan Taori, Tianyi Zhang, Ishaan Gulrajani, Jimmy Ba, Carlos Guestrin, Percy Liang, and Tatsunori~B. Hashimoto. 2023.
\newblock \href {http://papers.nips.cc/paper\_files/paper/2023/hash/5fc47800ee5b30b8777fdd30abcaaf3b-Abstract-Conference.html} {Alpacafarm: {A} simulation framework for methods that learn from human feedback}.
\newblock In \emph{Advances in Neural Information Processing Systems 36: Annual Conference on Neural Information Processing Systems 2023, NeurIPS 2023, New Orleans, LA, USA, December 10 - 16, 2023}.

\bibitem[{Hendrycks et~al.(2021{\natexlab{a}})Hendrycks, Burns, Basart, Zou, Mazeika, Song, and Steinhardt}]{DBLP:conf/iclr/HendrycksBBZMSS21}
Dan Hendrycks, Collin Burns, Steven Basart, Andy Zou, Mantas Mazeika, Dawn Song, and Jacob Steinhardt. 2021{\natexlab{a}}.
\newblock \href {https://openreview.net/forum?id=d7KBjmI3GmQ} {Measuring massive multitask language understanding}.
\newblock In \emph{9th International Conference on Learning Representations, {ICLR} 2021, Virtual Event, Austria, May 3-7, 2021}. OpenReview.net.

\bibitem[{Hendrycks et~al.(2021{\natexlab{b}})Hendrycks, Burns, Kadavath, Arora, Basart, Tang, Song, and Steinhardt}]{DBLP:conf/nips/HendrycksBKABTS21}
Dan Hendrycks, Collin Burns, Saurav Kadavath, Akul Arora, Steven Basart, Eric Tang, Dawn Song, and Jacob Steinhardt. 2021{\natexlab{b}}.
\newblock \href {https://datasets-benchmarks-proceedings.neurips.cc/paper/2021/hash/be83ab3ecd0db773eb2dc1b0a17836a1-Abstract-round2.html} {Measuring mathematical problem solving with the {MATH} dataset}.
\newblock In \emph{Proceedings of the Neural Information Processing Systems Track on Datasets and Benchmarks 1, NeurIPS Datasets and Benchmarks 2021, December 2021, virtual}.

\bibitem[{Hu et~al.(2022)Hu, Shen, Wallis, Allen{-}Zhu, Li, Wang, Wang, and Chen}]{DBLP:conf/iclr/HuSWALWWC22}
Edward~J. Hu, Yelong Shen, Phillip Wallis, Zeyuan Allen{-}Zhu, Yuanzhi Li, Shean Wang, Lu~Wang, and Weizhu Chen. 2022.
\newblock \href {https://openreview.net/forum?id=nZeVKeeFYf9} {Lora: Low-rank adaptation of large language models}.
\newblock In \emph{The Tenth International Conference on Learning Representations, {ICLR} 2022, Virtual Event, April 25-29, 2022}. OpenReview.net.

\bibitem[{Huang et~al.(2023)Huang, Bai, Zhu, Zhang, Zhang, Su, Liu, Lv, Zhang, Lei, Fu, Sun, and He}]{DBLP:conf/nips/HuangBZZZSLLZLF23}
Yuzhen Huang, Yuzhuo Bai, Zhihao Zhu, Junlei Zhang, Jinghan Zhang, Tangjun Su, Junteng Liu, Chuancheng Lv, Yikai Zhang, Jiayi Lei, Yao Fu, Maosong Sun, and Junxian He. 2023.
\newblock \href {http://papers.nips.cc/paper\_files/paper/2023/hash/c6ec1844bec96d6d32ae95ae694e23d8-Abstract-Datasets\_and\_Benchmarks.html} {C-eval: {A} multi-level multi-discipline chinese evaluation suite for foundation models}.
\newblock In \emph{Advances in Neural Information Processing Systems 36: Annual Conference on Neural Information Processing Systems 2023, NeurIPS 2023, New Orleans, LA, USA, December 10 - 16, 2023}.

\bibitem[{Jiang et~al.(2023)Jiang, Sablayrolles, Mensch, Bamford, Chaplot, de~Las~Casas, Bressand, Lengyel, Lample, Saulnier, Lavaud, Lachaux, Stock, Scao, Lavril, Wang, Lacroix, and Sayed}]{DBLP:journals/corr/abs-2310-06825}
Albert~Q. Jiang, Alexandre Sablayrolles, Arthur Mensch, Chris Bamford, Devendra~Singh Chaplot, Diego de~Las~Casas, Florian Bressand, Gianna Lengyel, Guillaume Lample, Lucile Saulnier, L{\'{e}}lio~Renard Lavaud, Marie{-}Anne Lachaux, Pierre Stock, Teven~Le Scao, Thibaut Lavril, Thomas Wang, Timoth{\'{e}}e Lacroix, and William~El Sayed. 2023.
\newblock \href {https://doi.org/10.48550/ARXIV.2310.06825} {Mistral 7b}.
\newblock \emph{CoRR}, abs/2310.06825.

\bibitem[{Jiang et~al.(2024)Jiang, Wang, Zeng, Zhong, Li, Mi, Shang, Jiang, Liu, and Wang}]{DBLP:conf/acl/Jiang0ZZLMS00W24}
Yuxin Jiang, Yufei Wang, Xingshan Zeng, Wanjun Zhong, Liangyou Li, Fei Mi, Lifeng Shang, Xin Jiang, Qun Liu, and Wei Wang. 2024.
\newblock \href {https://doi.org/10.18653/V1/2024.ACL-LONG.257} {Followbench: {A} multi-level fine-grained constraints following benchmark for large language models}.
\newblock In \emph{Proceedings of the 62nd Annual Meeting of the Association for Computational Linguistics (Volume 1: Long Papers), {ACL} 2024, Bangkok, Thailand, August 11-16, 2024}, pages 4667--4688. Association for Computational Linguistics.

\bibitem[{K{\"{o}}pf et~al.(2023)K{\"{o}}pf, Kilcher, von R{\"{u}}tte, Anagnostidis, Tam, Stevens, Barhoum, Nguyen, Stanley, Nagyfi, ES, Suri, Glushkov, Dantuluri, Maguire, Schuhmann, Nguyen, and Mattick}]{DBLP:conf/nips/KopfKRATSBNSNES23}
Andreas K{\"{o}}pf, Yannic Kilcher, Dimitri von R{\"{u}}tte, Sotiris Anagnostidis, Zhi~Rui Tam, Keith Stevens, Abdullah Barhoum, Duc Nguyen, Oliver Stanley, Rich{\'{a}}rd Nagyfi, Shahul ES, Sameer Suri, David Glushkov, Arnav Dantuluri, Andrew Maguire, Christoph Schuhmann, Huu Nguyen, and Alexander Mattick. 2023.
\newblock \href {http://papers.nips.cc/paper\_files/paper/2023/hash/949f0f8f32267d297c2d4e3ee10a2e7e-Abstract-Datasets\_and\_Benchmarks.html} {Openassistant conversations - democratizing large language model alignment}.
\newblock In \emph{Advances in Neural Information Processing Systems 36: Annual Conference on Neural Information Processing Systems 2023, NeurIPS 2023, New Orleans, LA, USA, December 10 - 16, 2023}.

\bibitem[{Kwon et~al.(2023)Kwon, Li, Zhuang, Sheng, Zheng, Yu, Gonzalez, Zhang, and Stoica}]{DBLP:conf/sosp/KwonLZ0ZY0ZS23}
Woosuk Kwon, Zhuohan Li, Siyuan Zhuang, Ying Sheng, Lianmin Zheng, Cody~Hao Yu, Joseph Gonzalez, Hao Zhang, and Ion Stoica. 2023.
\newblock \href {https://doi.org/10.1145/3600006.3613165} {Efficient memory management for large language model serving with pagedattention}.
\newblock In \emph{Proceedings of the 29th Symposium on Operating Systems Principles, {SOSP} 2023, Koblenz, Germany, October 23-26, 2023}, pages 611--626. {ACM}.

\bibitem[{Li et~al.(2024)Li, Zhang, Li, Chen, Chen, Cheng, Wang, Zhou, and Xiao}]{DBLP:conf/naacl/LiZLCC0W0024}
Ming Li, Yong Zhang, Zhitao Li, Jiuhai Chen, Lichang Chen, Ning Cheng, Jianzong Wang, Tianyi Zhou, and Jing Xiao. 2024.
\newblock \href {https://doi.org/10.18653/V1/2024.NAACL-LONG.421} {From quantity to quality: Boosting {LLM} performance with self-guided data selection for instruction tuning}.
\newblock In \emph{Proceedings of the 2024 Conference of the North American Chapter of the Association for Computational Linguistics: Human Language Technologies (Volume 1: Long Papers), {NAACL} 2024, Mexico City, Mexico, June 16-21, 2024}, pages 7602--7635. Association for Computational Linguistics.

\bibitem[{Liu et~al.(2024)Liu, Zeng, He, Jiang, and He}]{DBLP:conf/iclr/0131Z00H24}
Wei Liu, Weihao Zeng, Keqing He, Yong Jiang, and Junxian He. 2024.
\newblock \href {https://openreview.net/forum?id=BTKAeLqLMw} {What makes good data for alignment? {A} comprehensive study of automatic data selection in instruction tuning}.
\newblock In \emph{The Twelfth International Conference on Learning Representations, {ICLR} 2024, Vienna, Austria, May 7-11, 2024}. OpenReview.net.

\bibitem[{Luo et~al.(2023)Luo, Sun, Xu, Zhao, Lou, Tao, Geng, Lin, Chen, and Zhang}]{DBLP:journals/corr/abs-2308-09583}
Haipeng Luo, Qingfeng Sun, Can Xu, Pu~Zhao, Jianguang Lou, Chongyang Tao, Xiubo Geng, Qingwei Lin, Shifeng Chen, and Dongmei Zhang. 2023.
\newblock \href {https://doi.org/10.48550/ARXIV.2308.09583} {Wizardmath: Empowering mathematical reasoning for large language models via reinforced evol-instruct}.
\newblock \emph{CoRR}, abs/2308.09583.

\bibitem[{Luo et~al.(2024)Luo, Xu, Zhao, Sun, Geng, Hu, Tao, Ma, Lin, and Jiang}]{DBLP:conf/iclr/LuoX0SGHT0LJ24}
Ziyang Luo, Can Xu, Pu~Zhao, Qingfeng Sun, Xiubo Geng, Wenxiang Hu, Chongyang Tao, Jing Ma, Qingwei Lin, and Daxin Jiang. 2024.
\newblock \href {https://openreview.net/forum?id=UnUwSIgK5W} {Wizardcoder: Empowering code large language models with evol-instruct}.
\newblock In \emph{The Twelfth International Conference on Learning Representations, {ICLR} 2024, Vienna, Austria, May 7-11, 2024}. OpenReview.net.

\bibitem[{OpenAI(2023)}]{DBLP:journals/corr/abs-2303-08774}
OpenAI. 2023.
\newblock \href {https://doi.org/10.48550/ARXIV.2303.08774} {{GPT-4} technical report}.
\newblock \emph{CoRR}, abs/2303.08774.

\bibitem[{Ouyang et~al.(2022)Ouyang, Wu, Jiang, Almeida, Wainwright, Mishkin, Zhang, Agarwal, Slama, Ray, Schulman, Hilton, Kelton, Miller, Simens, Askell, Welinder, Christiano, Leike, and Lowe}]{DBLP:conf/nips/Ouyang0JAWMZASR22}
Long Ouyang, Jeffrey Wu, Xu~Jiang, Diogo Almeida, Carroll~L. Wainwright, Pamela Mishkin, Chong Zhang, Sandhini Agarwal, Katarina Slama, Alex Ray, John Schulman, Jacob Hilton, Fraser Kelton, Luke Miller, Maddie Simens, Amanda Askell, Peter Welinder, Paul~F. Christiano, Jan Leike, and Ryan Lowe. 2022.
\newblock \href {http://papers.nips.cc/paper\_files/paper/2022/hash/b1efde53be364a73914f58805a001731-Abstract-Conference.html} {Training language models to follow instructions with human feedback}.
\newblock In \emph{Advances in Neural Information Processing Systems 35: Annual Conference on Neural Information Processing Systems 2022, NeurIPS 2022, New Orleans, LA, USA, November 28 - December 9, 2022}.

\bibitem[{Song et~al.(2020)Song, Tan, Qin, Lu, and Liu}]{DBLP:conf/nips/Song0QLL20}
Kaitao Song, Xu~Tan, Tao Qin, Jianfeng Lu, and Tie{-}Yan Liu. 2020.
\newblock \href {https://proceedings.neurips.cc/paper/2020/hash/c3a690be93aa602ee2dc0ccab5b7b67e-Abstract.html} {Mpnet: Masked and permuted pre-training for language understanding}.
\newblock In \emph{Advances in Neural Information Processing Systems 33: Annual Conference on Neural Information Processing Systems 2020, NeurIPS 2020, December 6-12, 2020, virtual}.

\bibitem[{Taori et~al.(2023)Taori, Gulrajani, Zhang, Dubois, Li, Guestrin, Liang, and Hashimoto}]{alpaca}
Rohan Taori, Ishaan Gulrajani, Tianyi Zhang, Yann Dubois, Xuechen Li, Carlos Guestrin, Percy Liang, and Tatsunori~B. Hashimoto. 2023.
\newblock Stanford alpaca: An instruction-following llama model.
\newblock \url{https://github.com/tatsu-lab/stanford_alpaca}.

\bibitem[{Team(2024)}]{qwen2.5}
Qwen Team. 2024.
\newblock \href {https://qwenlm.github.io/blog/qwen2.5/} {Qwen2.5: A party of foundation models}.

\bibitem[{Wang et~al.(2023)Wang, Kordi, Mishra, Liu, Smith, Khashabi, and Hajishirzi}]{DBLP:conf/acl/WangKMLSKH23}
Yizhong Wang, Yeganeh Kordi, Swaroop Mishra, Alisa Liu, Noah~A. Smith, Daniel Khashabi, and Hannaneh Hajishirzi. 2023.
\newblock \href {https://doi.org/10.18653/V1/2023.ACL-LONG.754} {Self-instruct: Aligning language models with self-generated instructions}.
\newblock In \emph{Proceedings of the 61st Annual Meeting of the Association for Computational Linguistics (Volume 1: Long Papers), {ACL} 2023, Toronto, Canada, July 9-14, 2023}, pages 13484--13508. Association for Computational Linguistics.

\bibitem[{Xu et~al.(2024{\natexlab{a}})Xu, Sun, Zheng, Geng, Zhao, Feng, Tao, Lin, and Jiang}]{DBLP:conf/iclr/XuSZG0FTLJ24}
Can Xu, Qingfeng Sun, Kai Zheng, Xiubo Geng, Pu~Zhao, Jiazhan Feng, Chongyang Tao, Qingwei Lin, and Daxin Jiang. 2024{\natexlab{a}}.
\newblock \href {https://openreview.net/forum?id=CfXh93NDgH} {Wizardlm: Empowering large pre-trained language models to follow complex instructions}.
\newblock In \emph{The Twelfth International Conference on Learning Representations, {ICLR} 2024, Vienna, Austria, May 7-11, 2024}. OpenReview.net.

\bibitem[{Xu et~al.(2024{\natexlab{b}})Xu, Jiang, Niu, Deng, Poovendran, Choi, and Lin}]{DBLP:journals/corr/abs-2406-08464}
Zhangchen Xu, Fengqing Jiang, Luyao Niu, Yuntian Deng, Radha Poovendran, Yejin Choi, and Bill~Yuchen Lin. 2024{\natexlab{b}}.
\newblock \href {https://doi.org/10.48550/ARXIV.2406.08464} {Magpie: Alignment data synthesis from scratch by prompting aligned llms with nothing}.
\newblock \emph{CoRR}, abs/2406.08464.

\bibitem[{Xu et~al.(2024{\natexlab{c}})Xu, Jiang, Niu, Lin, and Poovendran}]{xu2024strongermodelsstrongerteachers}
Zhangchen Xu, Fengqing Jiang, Luyao Niu, Bill~Yuchen Lin, and Radha Poovendran. 2024{\natexlab{c}}.
\newblock \href {https://arxiv.org/abs/2411.07133} {Stronger models are not stronger teachers for instruction tuning}.
\newblock \emph{Preprint}, arXiv:2411.07133.

\bibitem[{Yang et~al.(2024)Yang, Yang, Hui, Zheng, Yu, Zhou, Li, Li, Liu, Huang, Dong, Wei, Lin, Tang, Wang, Yang, Tu, Zhang, Ma, Yang, Xu, Zhou, Bai, He, Lin, Dang, Lu, Chen, Yang, Li, Xue, Ni, Zhang, Wang, Peng, Men, Gao, Lin, Wang, Bai, Tan, Zhu, Li, Liu, Ge, Deng, Zhou, Ren, Zhang, Wei, Ren, Liu, Fan, Yao, Zhang, Wan, Chu, Liu, Cui, Zhang, Guo, and Fan}]{DBLP:journals/corr/abs-2407-10671}
An~Yang, Baosong Yang, Binyuan Hui, Bo~Zheng, Bowen Yu, Chang Zhou, Chengpeng Li, Chengyuan Li, Dayiheng Liu, Fei Huang, Guanting Dong, Haoran Wei, Huan Lin, Jialong Tang, Jialin Wang, Jian Yang, Jianhong Tu, Jianwei Zhang, Jianxin Ma, Jianxin Yang, Jin Xu, Jingren Zhou, Jinze Bai, Jinzheng He, Junyang Lin, Kai Dang, Keming Lu, Keqin Chen, Kexin Yang, Mei Li, Mingfeng Xue, Na~Ni, Pei Zhang, Peng Wang, Ru~Peng, Rui Men, Ruize Gao, Runji Lin, Shijie Wang, Shuai Bai, Sinan Tan, Tianhang Zhu, Tianhao Li, Tianyu Liu, Wenbin Ge, Xiaodong Deng, Xiaohuan Zhou, Xingzhang Ren, Xinyu Zhang, Xipin Wei, Xuancheng Ren, Xuejing Liu, Yang Fan, Yang Yao, Yichang Zhang, Yu~Wan, Yunfei Chu, Yuqiong Liu, Zeyu Cui, Zhenru Zhang, Zhifang Guo, and Zhihao Fan. 2024.
\newblock \href {https://doi.org/10.48550/ARXIV.2407.10671} {Qwen2 technical report}.
\newblock \emph{CoRR}, abs/2407.10671.

\bibitem[{Zeng et~al.(2024)Zeng, Xu, Zhao, Lou, and Chen}]{DBLP:conf/emnlp/ZengXZLC24}
Weihao Zeng, Can Xu, Yingxiu Zhao, Jian{-}Guang Lou, and Weizhu Chen. 2024.
\newblock \href {https://aclanthology.org/2024.emnlp-main.397} {Automatic instruction evolving for large language models}.
\newblock In \emph{Proceedings of the 2024 Conference on Empirical Methods in Natural Language Processing, {EMNLP} 2024, Miami, FL, USA, November 12-16, 2024}, pages 6998--7018. Association for Computational Linguistics.

\bibitem[{Zhang et~al.(2023)Zhang, Dong, Li, Zhang, Sun, Wang, Li, Hu, Zhang, Wu, and Wang}]{DBLP:journals/corr/abs-2308-10792}
Shengyu Zhang, Linfeng Dong, Xiaoya Li, Sen Zhang, Xiaofei Sun, Shuhe Wang, Jiwei Li, Runyi Hu, Tianwei Zhang, Fei Wu, and Guoyin Wang. 2023.
\newblock \href {https://doi.org/10.48550/ARXIV.2308.10792} {Instruction tuning for large language models: {A} survey}.
\newblock \emph{CoRR}, abs/2308.10792.

\bibitem[{Zhao et~al.(2024)Zhao, Ren, Hessel, Cardie, Choi, and Deng}]{DBLP:conf/iclr/Zhao0HC0D24}
Wenting Zhao, Xiang Ren, Jack Hessel, Claire Cardie, Yejin Choi, and Yuntian Deng. 2024.
\newblock \href {https://openreview.net/forum?id=Bl8u7ZRlbM} {Wildchat: 1m chatgpt interaction logs in the wild}.
\newblock In \emph{The Twelfth International Conference on Learning Representations, {ICLR} 2024, Vienna, Austria, May 7-11, 2024}. OpenReview.net.

\bibitem[{Zheng et~al.(2024{\natexlab{a}})Zheng, Chiang, Sheng, Li, Zhuang, Wu, Zhuang, Li, Lin, Xing, Gonzalez, Stoica, and Zhang}]{DBLP:conf/iclr/ZhengC0LZW00LXG24}
Lianmin Zheng, Wei{-}Lin Chiang, Ying Sheng, Tianle Li, Siyuan Zhuang, Zhanghao Wu, Yonghao Zhuang, Zhuohan Li, Zi~Lin, Eric~P. Xing, Joseph~E. Gonzalez, Ion Stoica, and Hao Zhang. 2024{\natexlab{a}}.
\newblock \href {https://openreview.net/forum?id=BOfDKxfwt0} {Lmsys-chat-1m: {A} large-scale real-world {LLM} conversation dataset}.
\newblock In \emph{The Twelfth International Conference on Learning Representations, {ICLR} 2024, Vienna, Austria, May 7-11, 2024}. OpenReview.net.

\bibitem[{Zheng et~al.(2024{\natexlab{b}})Zheng, Zhang, Zhang, Ye, Luo, and Ma}]{DBLP:journals/corr/abs-2403-13372}
Yaowei Zheng, Richong Zhang, Junhao Zhang, Yanhan Ye, Zheyan Luo, and Yongqiang Ma. 2024{\natexlab{b}}.
\newblock \href {https://doi.org/10.48550/ARXIV.2403.13372} {Llamafactory: Unified efficient fine-tuning of 100+ language models}.
\newblock \emph{CoRR}, abs/2403.13372.

\bibitem[{Zhou et~al.(2023{\natexlab{a}})Zhou, Liu, Xu, Iyer, Sun, Mao, Ma, Efrat, Yu, Yu, Zhang, Ghosh, Lewis, Zettlemoyer, and Levy}]{DBLP:conf/nips/ZhouLX0SMMEYYZG23}
Chunting Zhou, Pengfei Liu, Puxin Xu, Srinivasan Iyer, Jiao Sun, Yuning Mao, Xuezhe Ma, Avia Efrat, Ping Yu, Lili Yu, Susan Zhang, Gargi Ghosh, Mike Lewis, Luke Zettlemoyer, and Omer Levy. 2023{\natexlab{a}}.
\newblock \href {http://papers.nips.cc/paper\_files/paper/2023/hash/ac662d74829e4407ce1d126477f4a03a-Abstract-Conference.html} {{LIMA:} less is more for alignment}.
\newblock In \emph{Advances in Neural Information Processing Systems 36: Annual Conference on Neural Information Processing Systems 2023, NeurIPS 2023, New Orleans, LA, USA, December 10 - 16, 2023}.

\bibitem[{Zhou et~al.(2023{\natexlab{b}})Zhou, Lu, Mishra, Brahma, Basu, Luan, Zhou, and Hou}]{DBLP:journals/corr/abs-2311-07911}
Jeffrey Zhou, Tianjian Lu, Swaroop Mishra, Siddhartha Brahma, Sujoy Basu, Yi~Luan, Denny Zhou, and Le~Hou. 2023{\natexlab{b}}.
\newblock \href {https://doi.org/10.48550/ARXIV.2311.07911} {Instruction-following evaluation for large language models}.
\newblock \emph{CoRR}, abs/2311.07911.

\end{thebibliography}

\appendix
\section{Appendix}
\label{sec:appendix}

\subsection{Experimental Details}
\label{sec:more_details}

\paragraph{Evolution Details of Evol-Instruct.} As shown in Figure~\ref{fig:depth},~\ref{fig:method_list} and~\ref{fig:breadth}, the instruction evolution prompts we utilized are derived from \citep{DBLP:conf/iclr/XuSZG0FTLJ24, DBLP:conf/iclr/LuoX0SGHT0LJ24}, with minor modifications. For the Alpaca dataset, we employ four in-depth evolution methods: \textit{deepening}, \textit{concretizing}, \textit{adding constraints}, and \textit{adding reasoning steps}, in addition to one breadth-focused evolution method. However, for the GSM8K Train and Code Alpaca datasets, we exclude the breadth-focused method and use only the four in-depth methods. To ensure a fair comparison, we apply these evolution methods to each original instruction in a fixed sequence, rather than randomly selecting them as in the original Evol-Instruct. This strategy is designed to eliminate variations in the evolution of the same instruction, thereby reducing the potential for biased experimental conclusions. The results in Table\ref{tab:llama_result} and Table~\ref{tab:qwen_result} are obtained after one round of evolution of the seed instructions.

\paragraph{Evolution Details of AutoIF.} Following the approach in AutoIF, we typically employ Llama-3.1-8B-Instruct and Llama-3.1-70B-Instruct to carry out the Instruction Augmentation and Verification steps, generating 780 and 420 instructions, respectively. Due to the multiple verification steps required by AutoIF for filtering, the number of generated instructions varies. To ensure fairness, we randomly select 420 instructions from the 780 generated by the SLMs for comparison. These instructions are then concatenated with queries from ShareGPT to create a dataset of 6,720 instruction data for subsequent training.

\paragraph{Evolution Details of Auto Evol-Instruct.} We compare the performance of Llama-3.1-8B-Instruct and Llama-3.1-70B-Instruct in automatically designing evolutionary trajectories for evolving instructions. Using the prompt template from Auto Evol-Instruct~\citep{DBLP:conf/emnlp/ZengXZLC24} (refer to Figure~\ref{fig:auto_evol}), we prompt the models to design evolutionary trajectories and evolve instructions autonomously. To avoid introducing additional bias, we exclude the optimization stage from Auto Evol-Instruct. The experimental setup and evaluation benchmarks are consistent with those in Section~\ref{sec:evol_instruct}. Since the models occasionally fail to adhere to the specified output format, leading to instruction extraction errors, we perform random sampling on the larger sets of evolved instructions from both models to ensure consistent quantities of instruction data. We also use Qwen-2.5-72B-Instruct to generate the responses. Finally, for the Alpaca, GSM8K, and Code Alpaca datasets, we conduct automatic evolution and sampling, resulting in 40,483, 6,200, and 15,533 instruction data points, respectively.

\paragraph{Implementation Details}
For a fair comparison, all of our experiments maintain consistent data volumes. During the construction of the instruction data, we leverage the vLLM framework~\citep{DBLP:conf/sosp/KwonLZ0ZY0ZS23} for acceleration using a temperature of 0.7 and a top\_p value of 0.95. For training the models, we utilize the LLaMA-Factory framework~\citep{DBLP:journals/corr/abs-2403-13372} with a global batch size of 64, a cutoff length of 2048, and a learning rate of 2e-5, following a cosine learning rate schedule over 3 epochs. No checkpoint selection is performed; instead, all models are evaluated using the final saved checkpoint. All experiments are carried out on 8 $\times$ NVIDIA Tesla A100 GPUs.

\paragraph{Base Models.} In the Evol-Instruct scenario, we fine-tune Llama series models~\citep{DBLP:journals/corr/abs-2407-21783} including Llama-3.2-3B, Llama-3.1-8B, Llama-3-8B, DeepSeek-7B~\citep{DBLP:journals/corr/abs-2401-02954}, Mistral-7B-v0.3~\citep{DBLP:journals/corr/abs-2310-06825}, and InternLM-2-7B~\citep{DBLP:journals/corr/abs-2403-17297} models. In the AutoIF scenario, we also use Llama series models as in Evol-Instruct, as well as Qwen series~\citep{DBLP:journals/corr/abs-2407-10671} models including Qwen-2.5-7B and Qwen-2-7B, along with InternLM-2-7B. For Auto Evol-Instruct, we evaluate the performance of the Llama series models.

\paragraph{More Hyperparameter Details.}
\begin{table}[ht]
    \centering
    \resizebox{0.48\textwidth}{!}{%
        \begin{tabular}{lc}
            \toprule
            \textbf{Hyperparameter} & \textbf{Value} \\
            \midrule
            Learning Rate & $2 \times 10^{-5}$ \\
            Number of Epochs & 3 \\
            Number of Devices & 8 \\
            Per-device Batch Size & 1 \\
            Gradient Accumulation Steps & 8 \\
            Learning Rate Scheduler & cosine \\
            Warmup Ratio & 0.03 \\
            Max Sequence Length & 2048 \\
            \bottomrule
        \end{tabular}
    }
    \caption{Hyperparameters utilized in Evol-Instruct, AutoIF and Auto Evol-Instruct scenarios.}
    \label{tab:sft_hyper}
\end{table}
\begin{table}[ht]
    \centering
    \resizebox{0.48\textwidth}{!}{%
        \begin{tabular}{lc}
            \toprule
            \textbf{Hyperparameter} & \textbf{Value} \\
            \midrule
            \multicolumn{2}{c}{\textit{General Hyperparameters}} \\
            Number of Epochs & 2 \\
            Number of Devices & 8 \\
            Per-device Batch Size & 1 \\
            Gradient Accumulation Steps & 8 \\
            Learning Rate Scheduler & cosine \\
            Warmup Ratio & 0.03 \\
            Max Sequence Length & 2048 \\
            \midrule
            \multicolumn{2}{c}{\textit{LoRA Hyperparameters}} \\
            LoRA Rank & 8 \\
            LoRA Alpha & 8 \\
            LoRA Target & all module \\
            LoRA Dropout & 0.0 \\
            \midrule
            \multicolumn{2}{c}{\textit{Qwen-2.5-0.5B and 1.5B}} \\
            Learning Rate & $1 \times 10^{-5}$ \\
            \midrule
            \multicolumn{2}{c}{\textit{Qwen-2.5-3B and 7B}} \\
            Learning Rate & $7 \times 10^{-6}$ \\
            \midrule
            \multicolumn{2}{c}{\textit{Qwen-2.5-14B, 32B and 72B}} \\
            Learning Rate & $5 \times 10^{-5}$ \\
            \bottomrule
        \end{tabular}
    }
    \caption{Hyperparameters utilized for fine-tuning Qwen-2.5 series models.}
    \label{tab:qwen_hyper}
\end{table}
We provide the detailed hyperparameters for supervised fine-tuning in Table~\ref{tab:sft_hyper}. Except for IFEval and FollowBench, which are evaluated using their respective repositories, all other evaluations are conducted using the OpenCompass~\citep{2023opencompass} framework, and vLLM is adopted for inference acceleration throughout the evaluation process to enhance computational efficiency and expedite the assessment procedures.

\subsection{Detailed Information of Seed Datasets}
\label{sec:seed_data}
\begin{table}[ht]
    \centering
    \resizebox{0.48\textwidth}{!}{%
        \begin{tabular}{lcc}
            \toprule
            \multirow{2}{*}{}   &   \multicolumn{2}{c}{\textbf{Seed Data}} \\
            \cmidrule{2-3}
                &   \textbf{Dataset} &   \textbf{Datasize}    \\
            \midrule
            Instruction Following   &   Alpaca  &   51,983  \\
            Mathematical Reasoning  &   GSM8K Train &   7,473    \\
            Code Generation &   Code Alpaca &   20,022  \\
            \bottomrule
        \end{tabular}
    }
    \caption{Statistics of seed instruction data used in the Evol-Instruct and Auto-Evol-Instruct scenarios.}
    \label{tab:statistic}
\end{table}
In Evol-Instruction and Auto Evol-Instruct scenarios, we utilize the following seed datasets for instruction following, mathematical reasoning, and code generation: (1) Alpaca, a dataset that contains about 52K instruction following data points, (2) GSM8K Train, a dataset that includes nearly 7K high-quality, linguistically diverse grade school math word problems; and (3) Code Alpaca, a code generation dataset comprising approximately 20K samples. Table~\ref{tab:statistic} presents the statistical information of the seed datasets.

In the AutoIF scenario, we follow the setup described in the AutoIF paper, using the seed instructions provided by the authors and the queries from ShareGPT to construct the instructions.

\subsection{Detailed Information of Evaluations}
\label{sec:evaluations}
To evaluate the instruction following capabilities of our models, we employ several benchmarks, including IFEval and FollowBench. IFEval consists of 25 types of verifiable instructions across approximately 500 prompts, while FollowBench is a fine-grained, constraint-based instruction following a benchmark with five difficulty levels. It includes diverse open-ended instructions that require evaluation by strong LLMs. We report both strict and loose accuracy metrics at the prompt and instruction levels, and for FollowBench, we specifically report the Hard Satisfaction Rate (HSR).

In addition to instruction following benchmarks, we assess the models on other tasks. For mathematical reasoning, we use GSM8K and MATH. GSM8K consists of grade school math problems, while MATH presents more challenging mathematical problems. We report accuracy scores for both datasets. For code generation, we evaluate the models using HumanEval and MBPP, reporting the pass@1 metrics. We also evaluate our models on C-Eval, and MMLU to provide a comprehensive assessment of the models' capabilities across various domains.

\subsection{More Experimental Results}
\label{sec:more_result}

\begin{table*}[ht]
    \centering
    \small
        \begin{tabular}{lcccccccccc}
            \toprule
            \multirow{2}{*}{\textbf{Model}} & \multicolumn{4}{c}{\textbf{Instruction Following (IFEval)}} & & \multicolumn{2}{c}{\textbf{Math Reasoning}} & & \multicolumn{2}{c}{\textbf{Code Generation}} \\
            \cmidrule{2-11}
            & \textbf{Pr.(S)} & \textbf{In.(S)} & \textbf{Pr.(L)} & \textbf{In.(L)} & & \textbf{GSM8K} & \textbf{MATH} & & \textbf{HumanEval} & \textbf{MBPP} \\
            \midrule
            \multicolumn{11}{c}{\textit{Seed instruction data}} \\
            Mistral-7B-v0.3 & 17.01 & 26.86 & 19.04 & 29.14 & \vrule & 27.07 & 0.12 & \vrule & 10.20 & 8.80 \\
            DeepSeek-7B & 22.00 & 34.05 & 23.48 & 35.73 & \vrule & 44.05 & 0.56 & \vrule & 25.61 & 33.80 \\
            Llama-3.2-3B & 22.55 & 34.17 & 25.88 & 37.65 & \vrule & 46.40 & 0.56 & \vrule & 28.05 & 32.20 \\
            Llama-3-8B & 23.11 & 32.97 & 24.77 & 35.13 & \vrule & 53.68 & 0.22 & \vrule & 25.00 & 28.60 \\
            Llama-3.1-8B & 27.54 & 38.13 & 28.65 & 39.21 & \vrule & 56.41 & 7.56 & \vrule & 29.88 & 31.80 \\
            InternLM-2-7B & 32.72 & 45.08 & 35.30 & 48.08 & \vrule & 61.87 & 10.28 & \vrule & 42.07 & 40.00 \\
            \bottomrule
        \end{tabular}
    \caption{Results of seed instruction data.}
    \label{tab:seed_result}
\end{table*}
\paragraph{Seed Instruction Data Results.} Table~\ref{tab:seed_result} presents the experimental results for the seed instruction datasets used in Evol-Instruct and Auto Evol-Instruct scenarios. We observe that the performance of models trained on these seed data is suboptimal. We argue that the quality of these seed data is no longer adequate to further improve the performance of the current advanced base models.

\begin{table*}[h]
    \centering
    \small
        \begin{tabular}{lcccccccccc}
            \toprule
            \multirow{2}{*}{\textbf{Model}} & \multicolumn{4}{c}{\textbf{Instruction Following (IFEval)}} & & \multicolumn{2}{c}{\textbf{Math Reasoning}} & & \multicolumn{2}{c}{\textbf{Code Generation}} \\
            \cmidrule{2-11}
            & \textbf{Pr.(S)} & \textbf{In.(S)} & \textbf{Pr.(L)} & \textbf{In.(L)} & & \textbf{GSM8K} & \textbf{MATH} & & \textbf{HumanEval} & \textbf{MBPP} \\
            \midrule
            \multicolumn{11}{c}{\textit{Supervised Model: Llama-3.1-70B-Instruct}}  \\
            Iteration 1 & 33.83 & 46.28 & 36.41 & 49.28 & \vrule & 63.00 & 7.62 & \vrule & 43.90 & 36.20 \\
            Iteration 2 & 32.53 & 43.76 & 34.20 & 46.16 & \vrule & 64.59 & 10.04 & \vrule & 42.07 & 36.60 \\
            Iteration 3 & 35.12 & 47.36 & 36.97 & 49.28 & \vrule & 64.75 & 11.82 & \vrule & 43.29 & 37.20 \\
            \midrule
            \multicolumn{11}{c}{\textit{Supervised Model: Llama-3.1-8B-Instruct}} \\
            Iteration 1 & 35.49 & 47.00 & 39.56 & 50.72 & \vrule & 63.38 & 11.44 & \vrule & 48.17 & 37.60 \\
            Iteration 2 & 36.78 & 48.20 & 40.30 & 50.84 & \vrule & 64.82 & 11.48 & \vrule & 48.78 & 39.40 \\
            Iteration 3 & 33.09 & 44.72 & 36.41 & 48.32 & \vrule & 65.88 & 14.12 & \vrule & 44.51 & 40.80 \\
            \bottomrule
        \end{tabular}
    \caption{Detailed performance of different evolved iterations on Llama-3-8B refer to Figure~\ref{fig:iterations}.}
    \label{tab:iterations}
\end{table*}

\paragraph{Detailed Results of Multi-Iteration Evolution.} Table~\ref{tab:iterations} presents the detailed results of different evolved iterations which are referred to Figure~\ref{fig:iterations}.

\begin{table*}[ht]
    \centering
    \small
        \begin{tabular}{lcccccccccc}
            \toprule
            \multirow{2}{*}{\textbf{Model}} & \multicolumn{4}{c}{\textbf{Instruction Following (IFEval)}} & & \multicolumn{2}{c}{\textbf{Math Reasoning}} & & \multicolumn{2}{c}{\textbf{Code Generation}} \\
            \cmidrule{2-11}
            & \textbf{Pr.(S)} & \textbf{In.(S)} & \textbf{Pr.(L)} & \textbf{In.(L)} & & \textbf{GSM8K} & \textbf{MATH} & & \textbf{HumanEval} & \textbf{MBPP} \\
            \midrule
            \multicolumn{11}{c}{\textit{Supervised Model: Llama-3.1-70B-Instruct}}  \\
            Qwen-2.5-0.5B & 18.48 & 32.73 & 22.00 & 35.85 & \vrule & 40.26 & 16.32 & \vrule & 30.49 & 27.60 \\
            Qwen-2.5-1.5B & 28.84 & 42.67 & 31.98 & 46.04 & \vrule & 62.32 & 24.06 & \vrule & 50.00 & 43.20 \\
            Qwen-2.5-3B & 37.89 & 48.56 & 42.70 & 53.60 & \vrule & 76.12 & 26.44 & \vrule & 63.41 & 55.40 \\
            Qwen-2.5-7B & 46.21 & 56.83 & 50.64 & 60.79 & \vrule & 76.12 & 38.14 & \vrule & 70.73 & 61.60 \\
            Qwen-2.5-14B \texttt{(LoRA)} & 40.11 & 54.43 & 48.24 & 61.99 & \vrule & 87.79 & 49.94 & \vrule & 75.00 & 67.20 \\
            Qwen-2.5-32B \texttt{(LoRA)} & 42.88 & 57.31 & 51.20 & 64.15 & \vrule & 87.79 & 55.02 & \vrule & 80.49 & 71.20 \\
            Qwen-2.5-72B \texttt{(LoRA)} & 50.63 & 68.43 & 57.12 & 70.98 & \vrule & 91.05 & 58.83 & \vrule & 82.93 & 76.00 \\
            \midrule
            \multicolumn{11}{c}{\textit{Supervised Model: Llama-3.1-8B-Instruct}} \\
            Qwen-2.5-0.5B & 17.38 & 29.38 & 19.78 & 32.01 & \vrule & 40.71 & 16.26 & \vrule & 34.76 & 28.00 \\
            Qwen-2.5-1.5B & 28.47 & 41.73 & 31.98 & 44.96 & \vrule & 65.35 & 27.84 & \vrule & 52.44 & 49.94 \\
            Qwen-2.5-3B & 38.82 & 49.76 & 42.51 & 53.96 & \vrule & 76.57 & 30.92 & \vrule & 64.02 & 55.80 \\
            Qwen-2.5-7B & 47.32 & 58.39 & 51.39 & 62.35 & \vrule & 82.03 & 43.78 & \vrule & 71.95 & 61.80 \\
            Qwen-2.5-14B \texttt{(LoRA)} & 42.51 & 55.16 & 51.02 & 62.47 & \vrule & 88.17 & 52.22 & \vrule & 75.61 & 67.20 \\
            Qwen-2.5-32B \texttt{(LoRA)} & 45.84 & 58.75 & 54.71 & 66.31 & \vrule & 89.61 & 55.28 & \vrule & 81.71 & 73.20 \\
            Qwen-2.5-72B \texttt{(LoRA)} & 52.79 & 72.56 & 61.25 & 73.27 & \vrule & 91.36 & 60.75 & \vrule & 84.67 & 76.80 \\
            \bottomrule
        \end{tabular}
    \caption{Detailed performance among Qwen-2.5 series models refer to Figure~\ref{fig:scaling}.}
    \label{tab:scaling}
\end{table*}

\paragraph{Detailed Results of Scaling Experiments.} Table~\ref{tab:scaling} presents the detailed results of the model scaling experiment shown in Figure~\ref{fig:scaling}.

\begin{table*}[ht]
\small
\centering
\begin{tabular}{lcccc}
\toprule
\multirow{2}{*}{\textbf{Temperature}} & \textbf{HumanEval}                           & \textbf{MBPP}                            & \textbf{HumanEval}                           & \textbf{MBPP}                           \\ \cmidrule{2-5} 
                             & \multicolumn{2}{c}{\textit{Supervised Model: Llama-3.1-70B-Instruct}} & \multicolumn{2}{c}{\textit{Supervised Model: Llama-3.1-8B-Instruct}} \\ \midrule
greedy                       & 37.20                               & 33.40                           & \textbf{39.63}                                    & \textbf{36.40}                               \\
0.1                          & 36.59                                    & 36.40                                & \textbf{37.80}                                    & \textbf{37.60}                               \\
0.3                          & 38.41                                    & 35.20                                & \textbf{39.63}                                    & \textbf{37.80}                               \\
0.5                          & 35.98                                    & 33.40                                & \textbf{37.80}                                    & \textbf{35.80}                               \\
0.7                          & 35.98                                    & \textbf{36.00}                                & \textbf{39.02}                                    & 32.80                               \\
0.9                          & 34.76                                    & 33.00                                & \textbf{40.24}                                    & \textbf{35.80}                               \\ \bottomrule
\end{tabular}
\caption{Performance among different temperatures on Llama-3.2-3B under code generation scenario.}
\label{tab:temperature}
\end{table*}

\paragraph{The Impact of Temperatures.} To explore the impact of temperature on the evolutionary instruction data, we compare Llama-3.1-8B-Instruct and Llama-3.1-70B-Instruct under different temperatures. Specifically, we evolve the Code Alpaca data under greedy decoding (with a temperature of 0) and at five different temperatures ranging from 0.1 to 0.9, and uniformly use Qwen-2.5-72B-Instruct to generate the corresponding responses. As shown in Table~\ref{tab:temperature}, the results of training on Llama-3.2-3B indicate that the SLMs perform consistently better than LLMs under all temperatures, which further validates the universality of our conclusion. 

\begin{figure}[ht]
    \centering
    \includegraphics[width=0.48\textwidth]{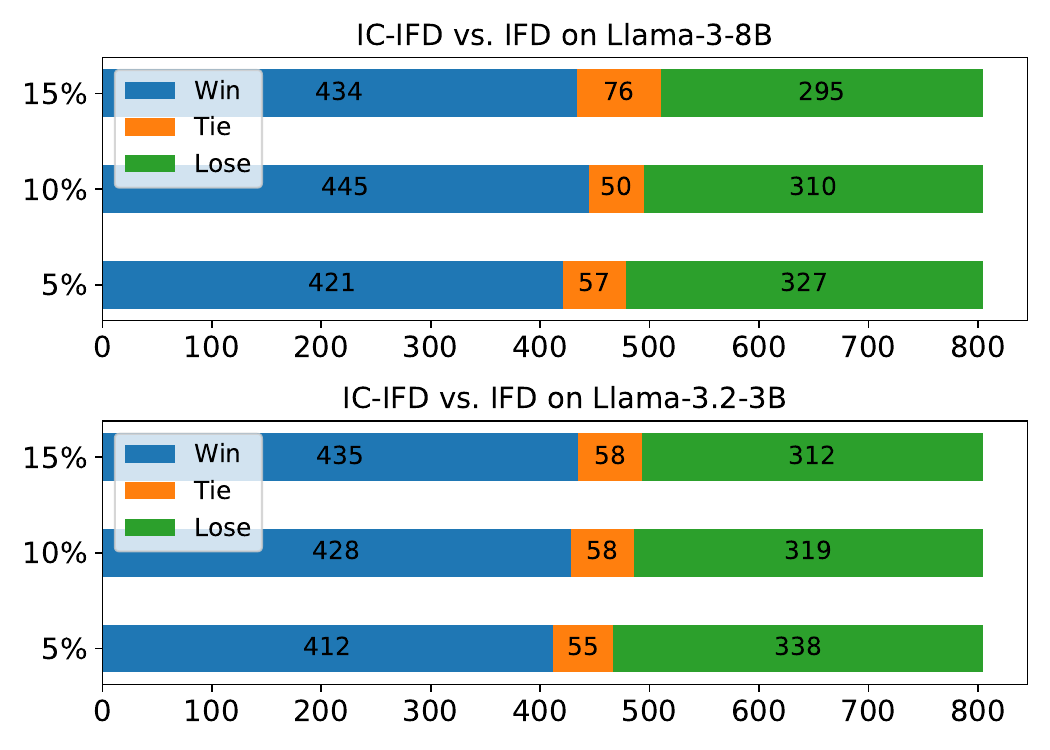}
    \caption{Performance comparison of three data selection ratios on Alpaca dataset between IC-IFD and IFD.}
    \label{fig:win} 
\end{figure}

\begin{figure}[ht]
    \centering
    \includegraphics[width=0.48\textwidth]{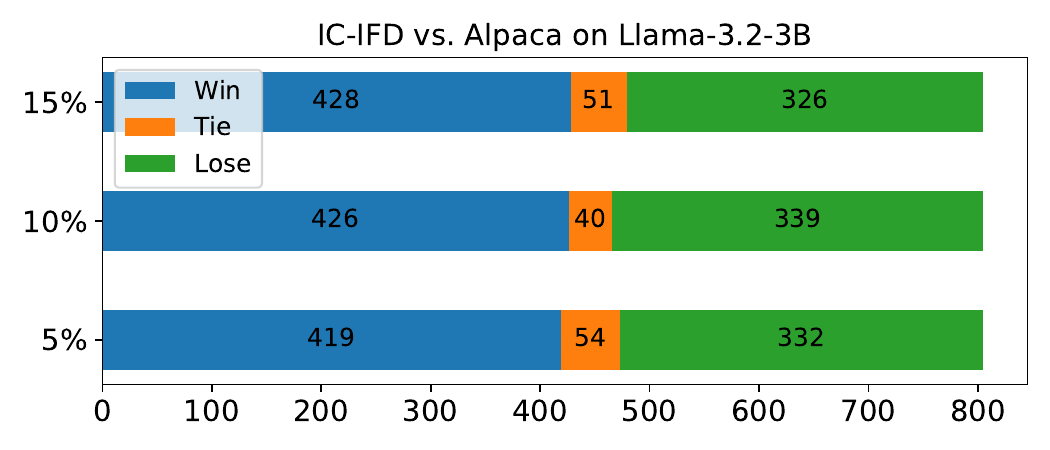}
    \caption{Performance comparison of three data selection ratios on Alpaca dataset between IC-IFD and full dataset.}
    \label{fig:win_all} 
\end{figure}

\paragraph{More Results of IC-IFD.} To further validate the broad applicability of IC-IFD, beyond high-difficulty instruction data, we use the IC-IFD and IFD metrics to filter 5\%, 10\%, and 15\% of the original Alpaca dataset for training the Llama-3-8B and Llama-3.2-3B models. We fine-tune the models on the IC-IFD and IFD-filtered data and evaluate their performance using instructions from AlpacaFarm~\citep{DBLP:conf/nips/DuboisLTZGBGLH23}. The generated responses are then assessed using GPT-4 to determine the win-tie-lose ratio (the evaluation prompt template can be found in Figure~\ref{fig:evaluation}). As shown in Figure~\ref{fig:win}, we observe that IC-IFD consistently outperforms IFD across all three data ratio settings for both models. Furthermore, we compare the performance of models trained on IC-IFD-filtered data with those trained on the full Alpaca dataset. As shown in Figure~\ref{fig:win_all}, models trained on IC-IFD-filtered data also perform better than those trained on the full dataset, further demonstrating the effectiveness of the proposed IC-IFD.

\begin{figure*}[ht] 
    \centering
    \includegraphics[width=\textwidth]{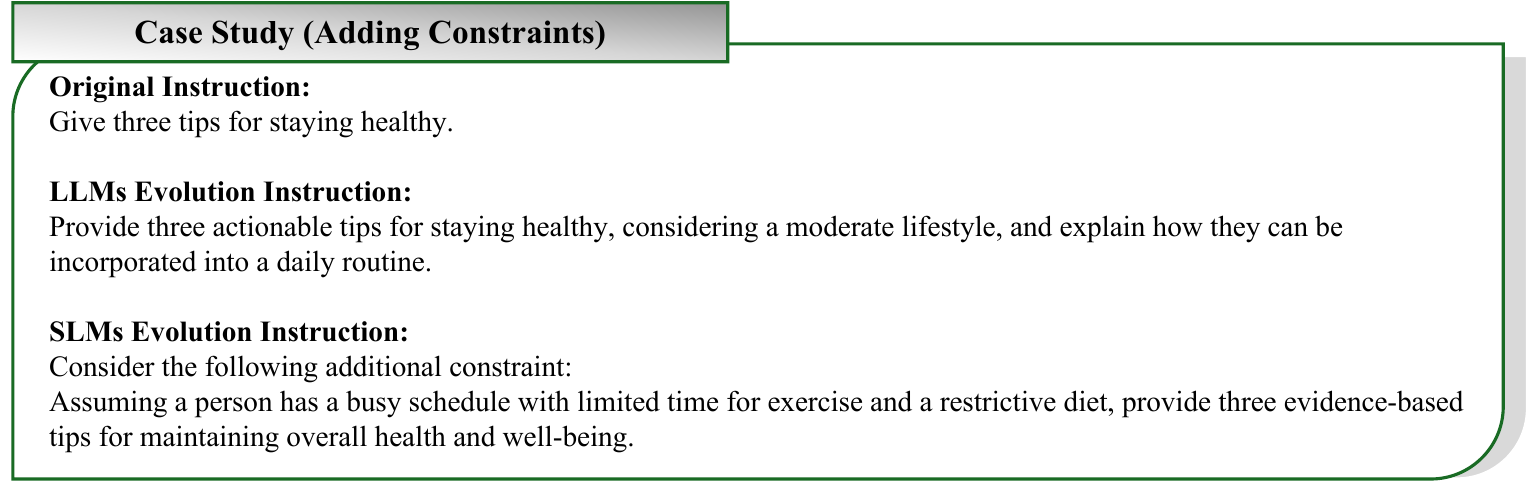}
    \caption{Comparison of cases between LLMs and SLMs under adding constraints strategy.} 
    \label{fig:constraint}
\end{figure*}

\begin{figure*}[ht] 
    \centering
    \includegraphics[width=\textwidth]{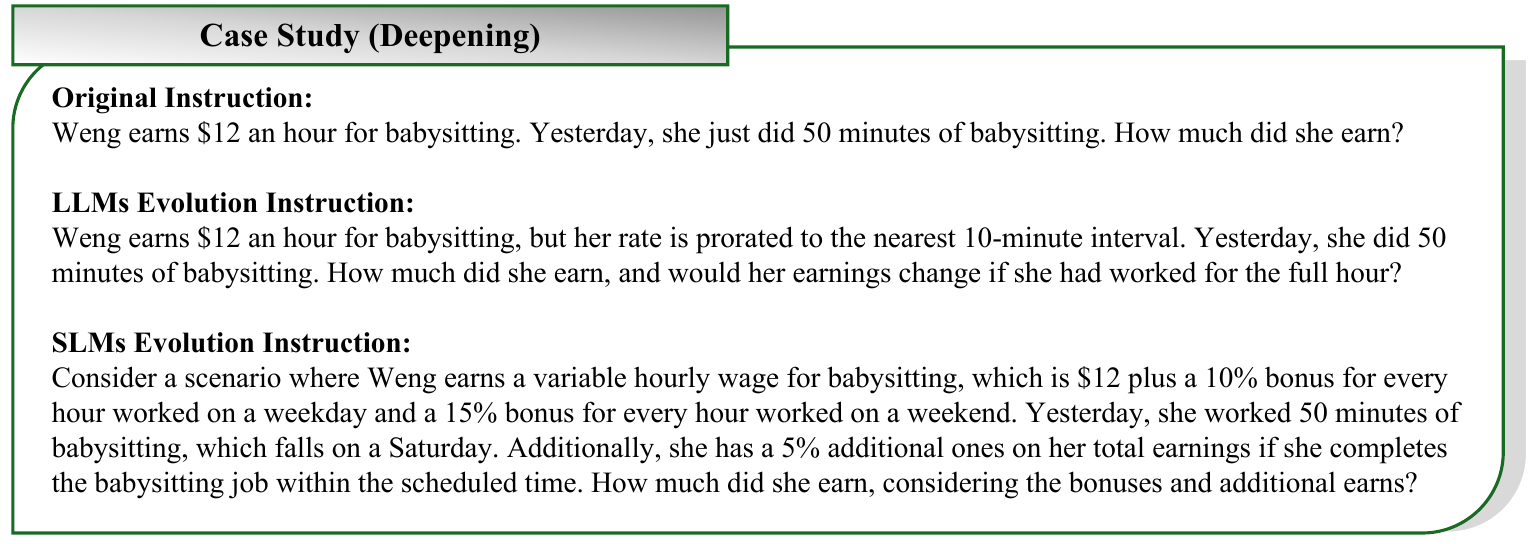}
    \caption{Comparison of cases between LLMs and SLMs under deepening strategy.} 
    \label{fig:deepening}
\end{figure*}

\paragraph{Case Study.} We compare the evolution of SLMs and LLMs across two specific in-depth cases. As illustrated in Figure~\ref{fig:constraint}, we observe that in the "adding constraints" evolution trajectory, the evolved instructions of SLMs incorporate two additional constraints: lack of time for exercise and inability to limit diet, while the evolved instructions of LLMs only add the condition that the requirements must be feasible. Similarly, in the "deepening" evolution trajectory, as shown in Figure~\ref{fig:deepening}, the evolved instructions of SLMs are significantly more challenging, containing numerous in-depth conditions, which is absent in the evolved instructions of LLMs. Overall, from actual cases, SLMs can evolve more complex and diverse instructions under the same constraints or trajectories, achieving more effective instructions at a lower computational cost.

\begin{table}[ht]
    \centering
    \resizebox{0.48\textwidth}{!}{%
        \begin{tabular}{lccc}
            \toprule
            & \textbf{Alpaca} & \textbf{GSM8K Train} & \textbf{Code Alpaca} \\
            \midrule
            Seed Instruction & 27.63 & 34.05 & 26.01 \\
            \midrule
            \textsc{LLM-Inst} Iter1 & 52.89 & 39.88 & 46.75 \\
            \textsc{SLM-Inst} Iter1 & \textbf{66.35} & \textbf{48.85} & \textbf{58.86} \\
            \midrule
            \textsc{LLM-Inst} Iter2 & 68.16 & 47.14 & 65.02 \\
            \textsc{SLM-Inst} Iter2 & \textbf{77.62} & \textbf{63.48} & \textbf{73.37} \\
            \midrule
            \textsc{LLM-Inst} Iter3 & 75.73 & 54.00 & 72.85 \\
            \textsc{SLM-Inst} Iter3 & \textbf{82.44} & \textbf{72.12} & \textbf{79.19} \\
            \bottomrule
        \end{tabular}
    }
    \caption{Scores of difficulty levels for instructions evolved during three iterations, using Llama-3.1-8B-Instruct and Llama-3.1-70B-Instruct as supervised models for each round under Evol-Instruct scenario.}
    \label{tab:difficulty}
\end{table}

\subsection{Further Analysis}
\label{sec:analysis}

\paragraph{Difficulty Scores of Evol-Instruct.} We utilize the prompt template shown in Figure~\ref{fig:score} to prompt Qwen-2.5-72B-Instruct for evaluating the complexity scores of the three-round data in the Evol-Instruct scenario. As shown in Table~\ref{tab:difficulty}, we find that in each round, \textsc{SLM-Inst} consistently outperforms \textsc{LLM-Inst} in terms of complexity scores. Interestingly, \textsc{SLM-Inst} Iter 2 is even more difficult than \textsc{LLM-Inst} Iter 3, as demonstrated by the experiment in Figure~\ref{fig:iterations}, where the overall performance of \textsc{SLM-Inst} Iter 2 is superior to that of \textsc{LLM-Inst} Iter 3.

\begin{table}[ht]
\resizebox{0.48\textwidth}{!}{%
\begin{tabular}{lccc}
\toprule
\multirow{2}{*}{\textbf{Iteration}} & \multicolumn{3}{c}{\textbf{Average Reward}}                     \\
\cmidrule{2-4}
                                    & \textbf{Alpaca} & \textbf{GSM8K} & \textbf{Code Alpaca} \\
\midrule
\multicolumn{4}{c}{\textit{Supervised Model: Llama-3.1-70B-Instruct}}                                  \\
Iteration 1                         & 1.54            & 0.74           & 1.10                 \\
Iteration 2                         & \textbf{1.68}   & 0.73           & \textbf{1.19}        \\
Iteration 3                         & \textbf{1.56}   & 0.69           & \textbf{1.14}        \\
\midrule
\multicolumn{4}{c}{\textit{Supervised Model: Llama-3.1-8B-Instruct}}                                   \\
Iteration 1                         & \textbf{1.59}   & \textbf{1.01}  & \textbf{1.23}        \\
Iteration 2                         & 1.54            & \textbf{0.79}  & 0.96                 \\
Iteration 3                         & 1.42            & \textbf{0.97}  & 1.03                 \\
\bottomrule
\end{tabular}
}
\caption{Comparison of average rewards among different iteration evolution instruction data.}
\label{tab:reward}
\end{table}

\paragraph{Quality Score Evaluated by Reward Model.} We also utilize InternLM-2-7B-Reward as the reward model to evaluate the average scores of the evolved instructions of both SLMs and LLMs. Specifically, given the evolved prompt templates (as shown in Figure~\ref{fig:depth} and~\ref{fig:breadth}), we then use the reward model to evaluate the rewards of the evolved instructions generated by SLMs and LLMs respectively and obtain the mean reward of the instruction set. As shown in Table~\ref{tab:reward}, we find that the overall scores of the instructions evaluated by the reward model are approximately in line with its performance during the training stage. However, on some datasets, it could not accurately reflect the quality of the instructions. Moreover, using the reward model cannot directly assess the quality of the instructions. Instead, it requires the meta-instructions used when constructing the instructions. Therefore, the reward model cannot be well applied to the evaluation of instructions. 

\begin{table}[ht]
\centering
\resizebox{0.48\textwidth}{!}{%
\begin{tabular}{lccc}
\toprule
\textbf{Datasets}             & \textbf{IFD (\%)}       & \textbf{IC-IFD (\%)}    & \textbf{Performance}    \\ \midrule
SLMs (Alpaca iter 3) & \textbf{83.04} & 35.89          & 40.64          \\
LLMs (Alpaca iter 3) & 82.03          & \textbf{37.05} & \textbf{42.18} \\ \bottomrule
\end{tabular}
}
\caption{Comparison of IFD and IC-IFD on third-round evolved Alpaca datasets from SLMs and LLMs.}
\label{tab:ifd}
\end{table}

\paragraph{Comparison of IFD and IC-IFD.} We analyze the third-round evolved Alpaca dataset for both SLMs and LLMs. Specifically, we compute the IFD and IC-IFD scores for each sample in both datasets and compare their average scores. As shown in Table~\ref{tab:ifd}, we evaluate the average performance of IFEval on the two datasets using Llama-3-8B. We find that when the instruction difficulty level is too high, the IFD score tends to increase. However, the performance of the fine-tuned models does not align with expectations. In contrast, the IC-IFD score effectively captures the influence of instruction complexity, offering a more accurate data quality assessment.

\subsection{Prompt Templates}
\label{sec:prompt}

\begin{figure*}[ht] 
    \centering
    \includegraphics[width=\textwidth]{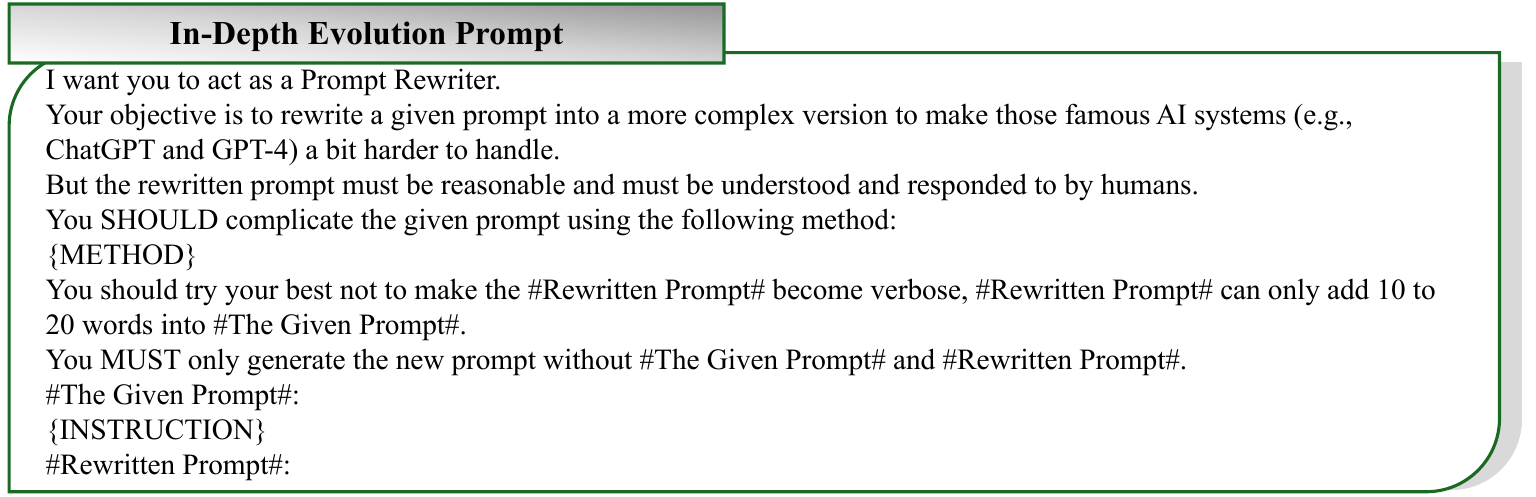}
    \caption{In-depth evolution prompt template utilized in Evol-Instruct scenario.} 
    \label{fig:depth}
\end{figure*}

\begin{figure*}[ht] 
    \centering
    \includegraphics[width=\textwidth]{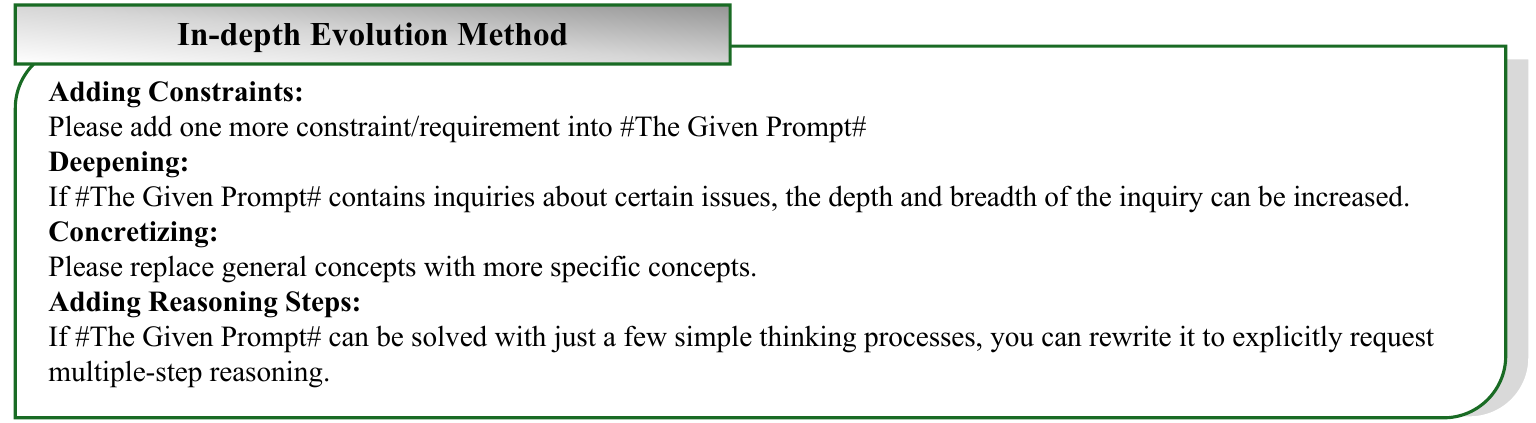}
    \caption{Four in-depth methods utilized in Evol-Instruct scenario.} 
    \label{fig:method_list}
\end{figure*}

\begin{figure*}[ht] 
    \centering
    \includegraphics[width=\textwidth]{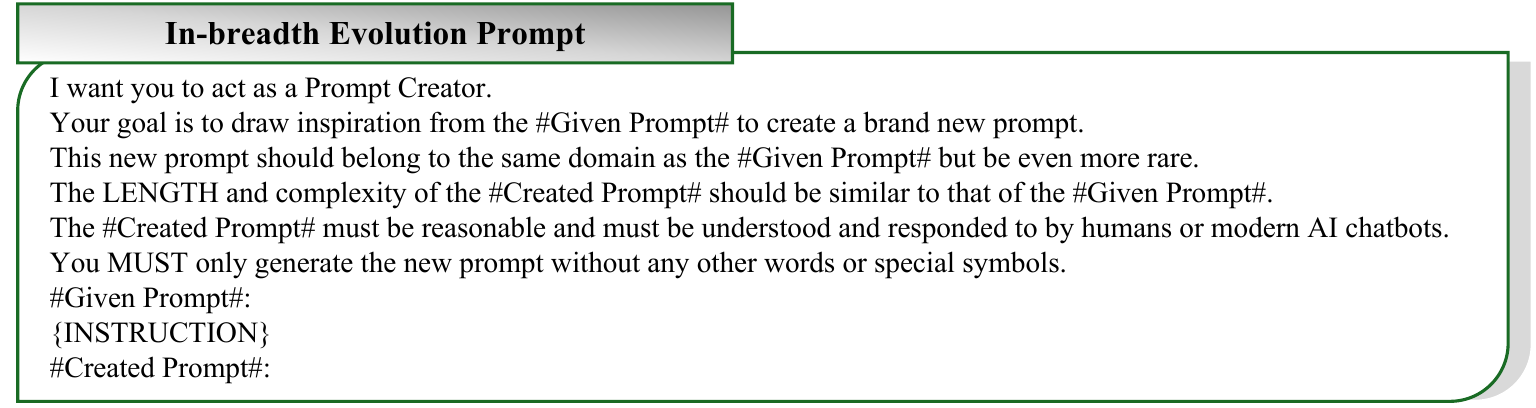}
    \caption{In-breadth evolution prompt template utilized in Evol-Instruct scenario.} 
    \label{fig:breadth}
\end{figure*}

\paragraph{Prompt Templates of Evol-Instruct.} Figure~\ref{fig:depth} shows the in-depth evolution prompt template for instruction evolution used in the Evol-Instruct scenario, derived from~\citep{DBLP:conf/iclr/XuSZG0FTLJ24} and slightly modified. Figures~\ref{fig:method_list} and~\ref{fig:breadth} demonstrate the four in-depth methods and one in-breadth evolved prompt template we adopt.

\begin{figure*}[ht] 
    \centering
    \includegraphics[width=\textwidth]{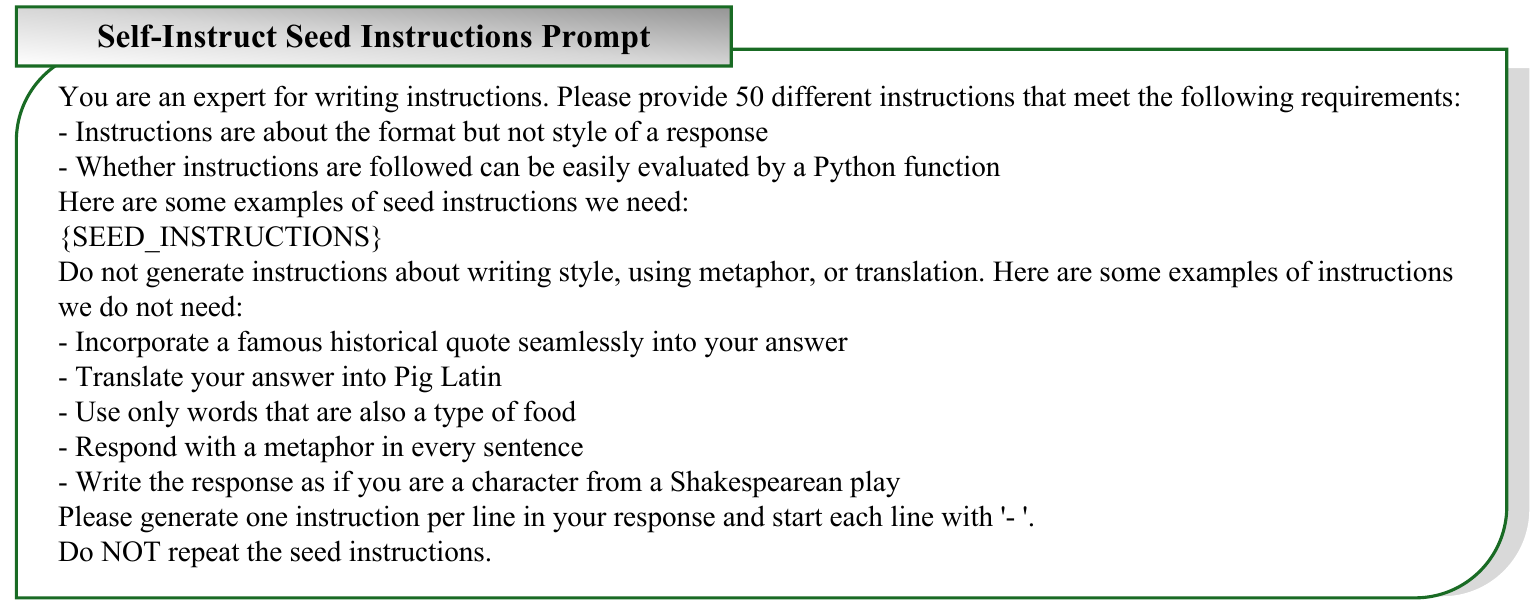}
    \caption{Prompt template of Self-Instruct Seed Instructions in AutoIF scenario.} 
    \label{fig:self_instruct}
\end{figure*}

\begin{figure*}[ht] 
    \centering
    \includegraphics[width=\textwidth]{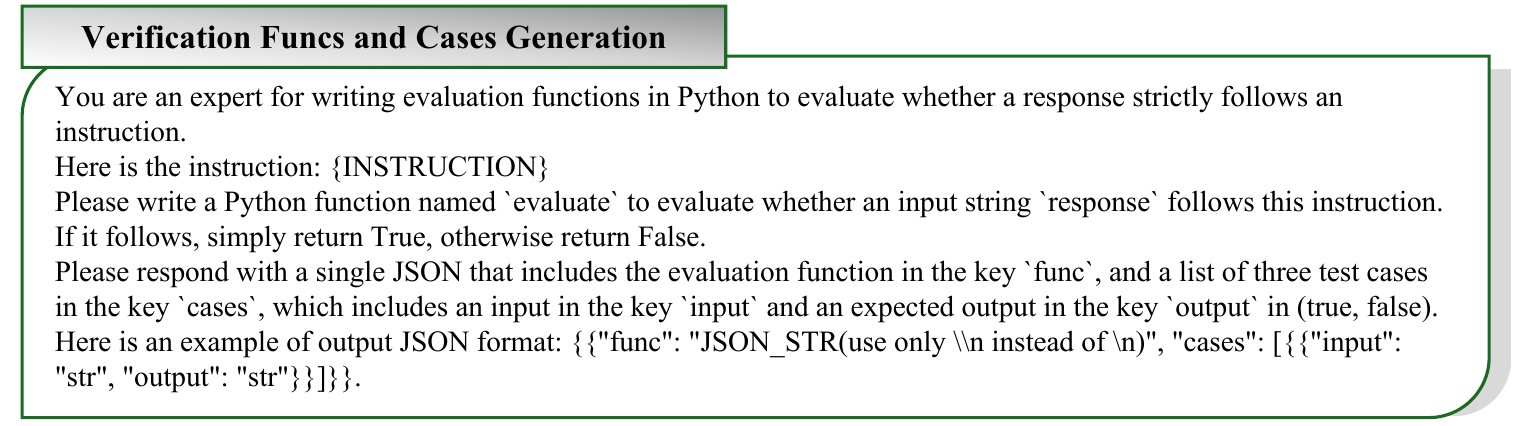}
    \caption{Prompt template of Verification Funcs and Cases Generation in AutoIF scenario.} 
    \label{fig:func_gen}
\end{figure*}

\paragraph{Prompt Templates of AutoIF.} We utilize the prompt templates consistent with those in~\citep{DBLP:journals/corr/abs-2406-13542}. Figures~\ref{fig:self_instruct} and~\ref{fig:func_gen} represent the prompts used in the two stages: Self-Instruct Seed Instructions and Verification Funcs and Cases Generation.

\begin{figure*}[ht] 
    \centering
    \includegraphics[width=\textwidth]{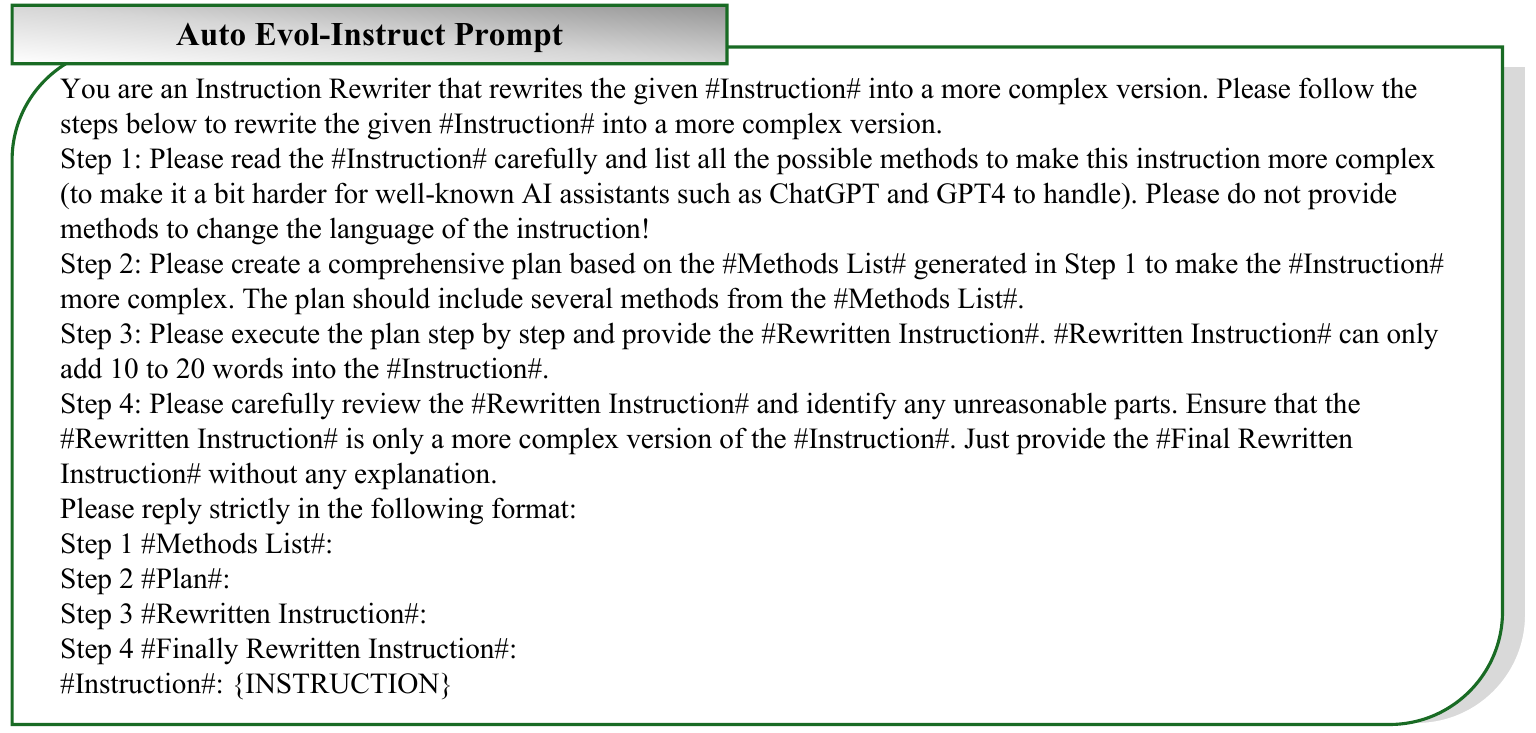}
    \caption{Prompt template of Auto Evol-Instruct scenario.} 
    \label{fig:auto_evol}
\end{figure*}

\paragraph{Prompt Templates of Auto Evol-Instruct.} As shown in Figure~\ref{fig:auto_evol}, we utilize the prompt templates consistent with those in~\citep{DBLP:conf/emnlp/ZengXZLC24} under Auto Evol-Instruct scenario.

\begin{figure*}[ht] 
    \centering
    \includegraphics[width=\textwidth]{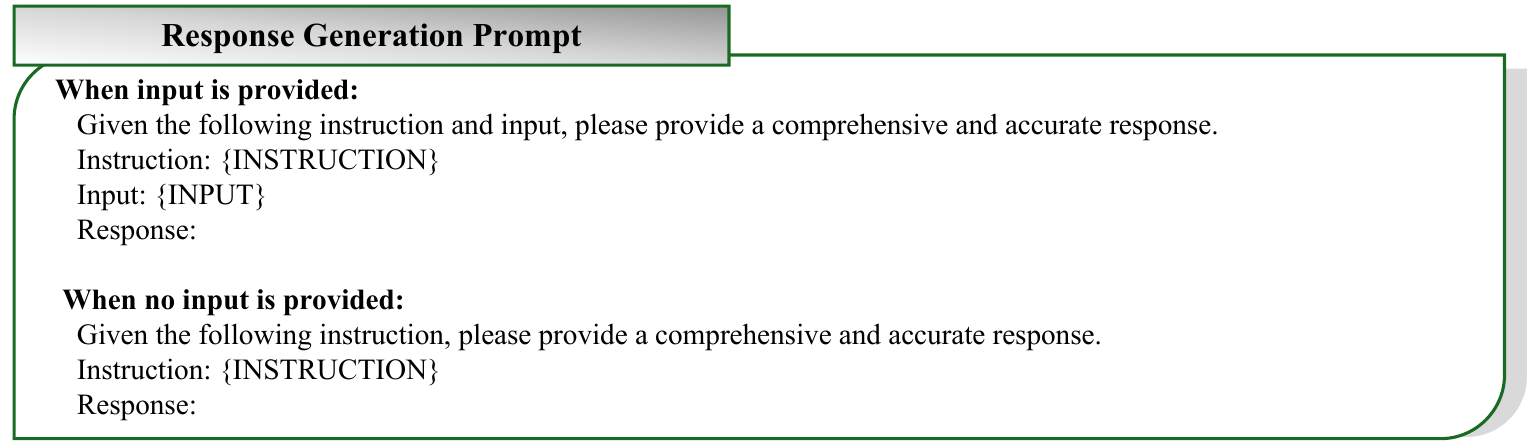}
    \caption{Prompt template of response generation.} 
    \label{fig:response}
\end{figure*}

\paragraph{Prompt Templates of Response Generation.} We use the prompt template shown in Figure~\ref{fig:response} to generate the corresponding responses for all instructions. We adopt the data organization format from Llama-Factory, and therefore, when generating responses, we classify them into two types based on the presence of an input.

\begin{figure*}[ht] 
    \centering
    \includegraphics[width=\textwidth]{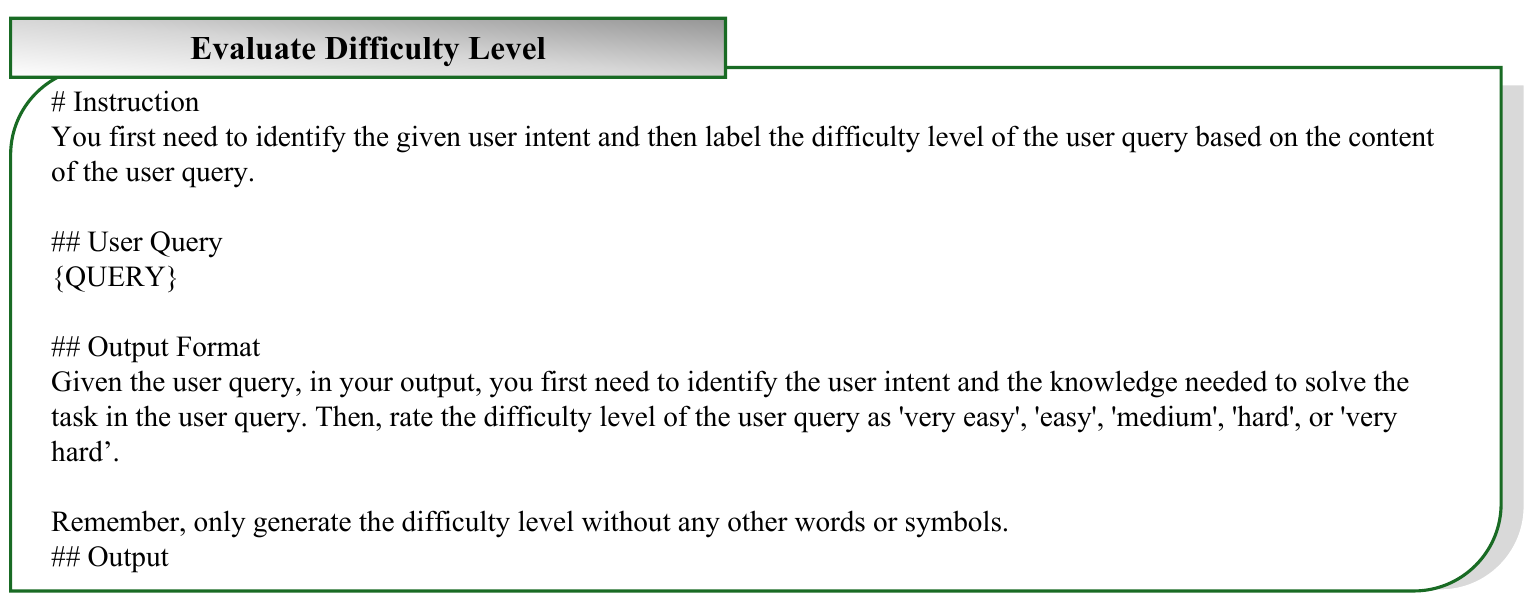}
    \caption{Prompt template of evaluating the difficulty levels.} 
    \label{fig:level}
\end{figure*}

\begin{figure*}[ht] 
    \centering
    \includegraphics[width=\textwidth]{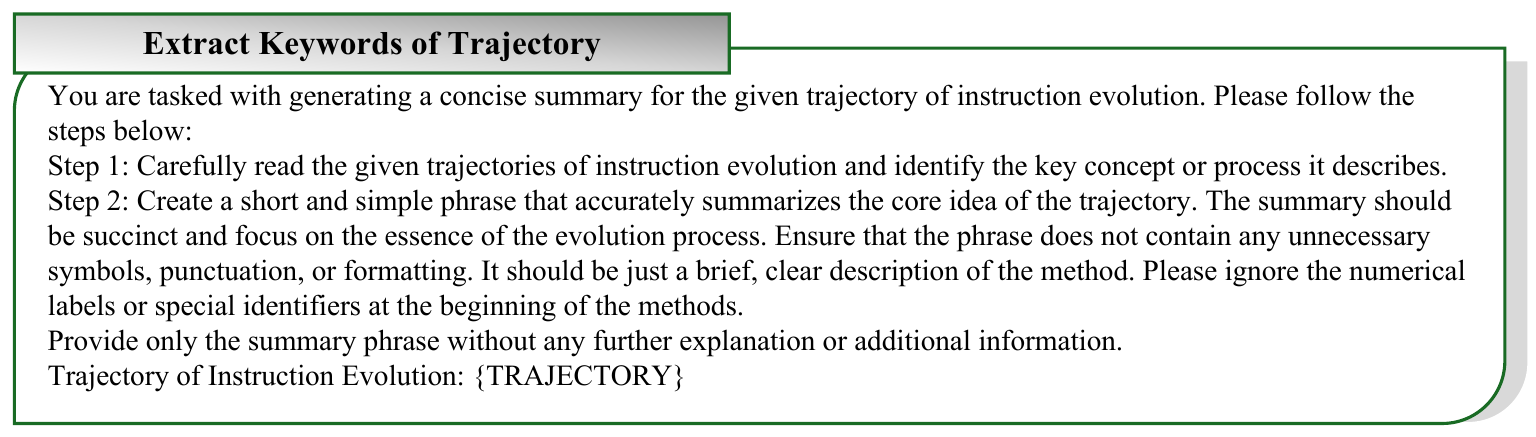}
    \caption{Prompt template of extracting the keywords from evolution trajectories.} 
    \label{fig:extract}
\end{figure*}

\begin{figure*}[ht] 
    \centering
    \includegraphics[width=\textwidth]{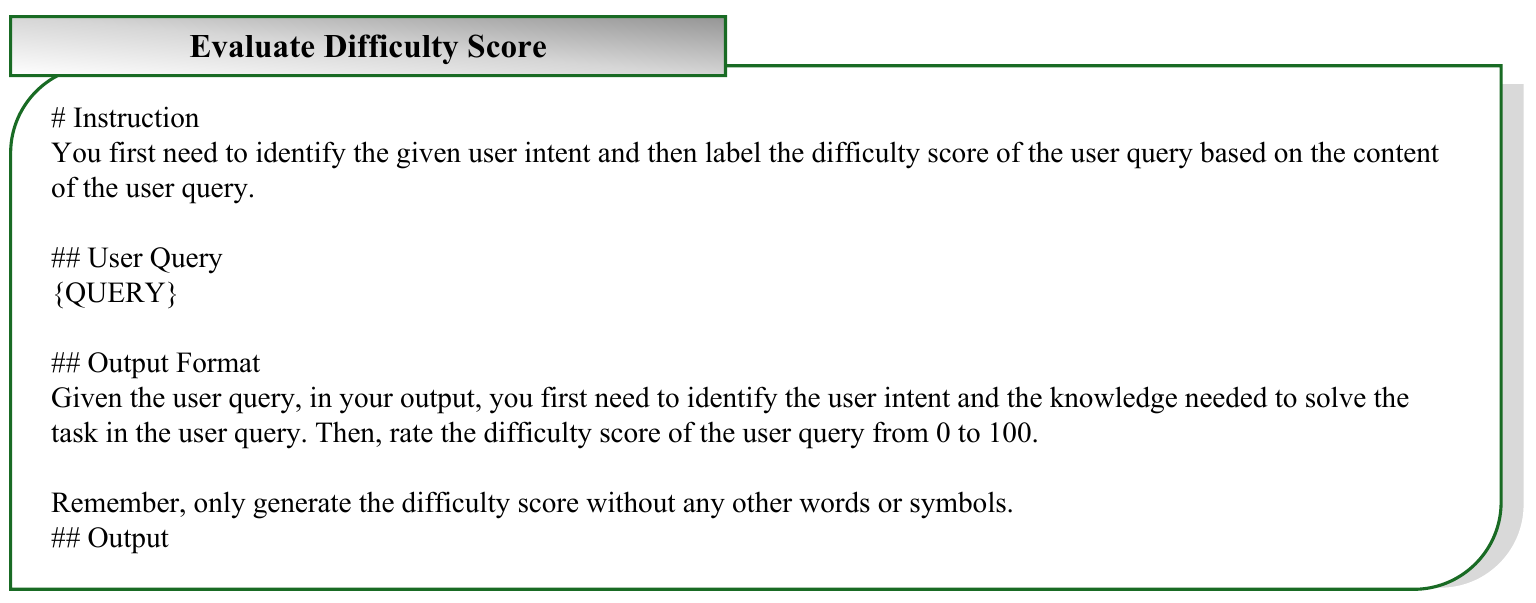}
    \caption{Prompt template of evaluating the difficulty scores.} 
    \label{fig:score}
\end{figure*}

\paragraph{Prompt Templates of Data Analysis.} Figure~\ref{fig:level} and~\ref{fig:score} show the prompt templates used to assess the difficulty levels and scores of instructions. Figure~\ref{fig:extract} displays the prompt template used to analyze the evolutionary trajectories automatically generated by the model.

\begin{figure*}[ht] 
    \centering
    \includegraphics[width=\textwidth]{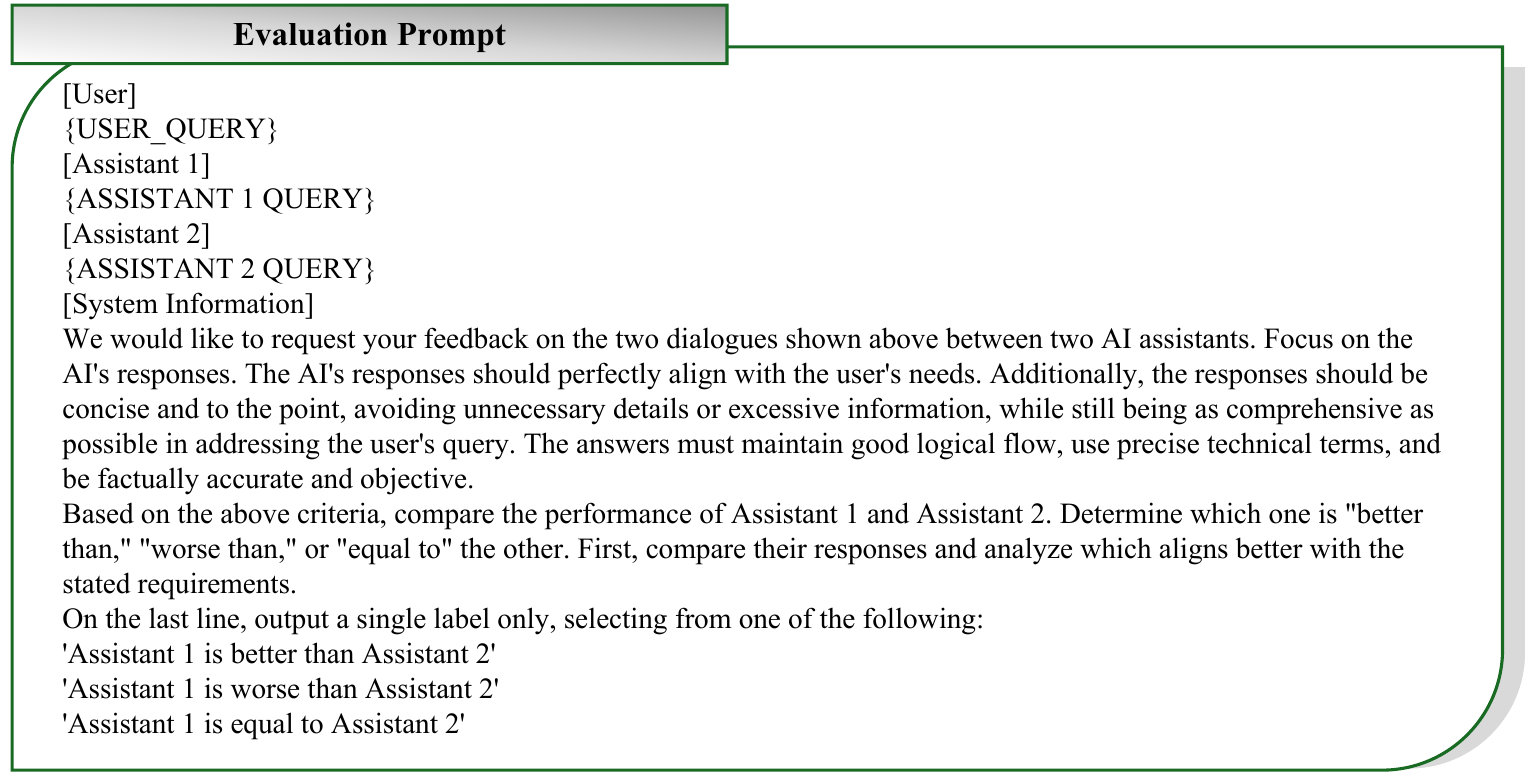}
    \caption{Prompt template of evaluating the win-tie-lose rates.} 
    \label{fig:evaluation}
\end{figure*}

\paragraph{Prompt Templates of Evaluation.} Figure~\ref{fig:evaluation} shows the prompt template used to assess the win-tie-lose rates on AlpacaFarm.

\end{document}